\documentclass[lettersize,journal]{IEEEtran}
\usepackage{amsmath,amsfonts}
\usepackage{algorithmic}
\usepackage{array}
\usepackage[caption=false,font=normalsize,labelfont=sf,textfont=sf]{subfig}
\usepackage{stfloats}
\usepackage{url}
\usepackage{verbatim}
\usepackage{graphicx}
\usepackage{multirow}
\usepackage{xcolor}
\usepackage{colortbl}
\usepackage{stix,bbding,pifont,utfsym,fontawesome}
\usepackage{booktabs}
\usepackage{cite}
\usepackage[pagebackref,breaklinks,colorlinks]{hyperref}
\def\BibTeX{{\rm B\kern-.05em{\sc i\kern-.025em b}\kern-.08em
    T\kern-.1667em\lower.7ex\hbox{E}\kern-.125emX}}
\usepackage{balance}

\definecolor{darkgreen}{HTML}{008000}

\begin{document}
\title{MOST: Motion Diffusion Model for Rare Text via Temporal Clip Banzhaf Interaction}
\author{Yin Wang, Mu li, Zhiying Leng, Frederick W. B. Li and Xiaohui Liang 
\thanks{Manuscript created March, 2024. This work was supported by the National Natural Science Foundation of China (Project Number: 62272019).
(Corresponding author: Xiaohui Liang) 

Yin Wang, Mu li and Zhiying Leng are with the State Key Laboratory of Virtual Reality Technology and Systems, Beihang University, Beijing, China (e-mail:wang\_yin, limu, zhiyingleng@buaa.edu.cn). 

Frederick W. B. Li is with the Department of Computer Science, University of Durham, Durham, UK (e-mail: frederick.li@durham.ac.uk).

Xiaohui Liang is with the State Key Laboratory of Virtual Reality Technology and Systems, Beihang University, Beijing, China, and also with Zhongguancun
Laboratory, Beijing, China (e-mail: liang\_xiaohui@buaa.edu.cn).}}

\markboth{Journal of \LaTeX\ Class Files,~Vol.~18, No.~9, September~2020}%
{How to Use the IEEEtran \LaTeX \ Templates}

\maketitle

\begin{abstract} 
We introduce \textbf{MOST}, a novel \textbf{MO}tion diffu\textbf{S}ion model via \textbf{T}emporal clip Banzhaf interaction, aimed at addressing the persistent challenge of generating human motion from rare language prompts. While previous approaches struggle with coarse-grained matching and overlook important semantic cues due to motion redundancy, our key insight lies in leveraging fine-grained clip relationships to mitigate these issues. MOST's retrieval stage presents the first formulation of its kind - temporal clip Banzhaf interaction - which precisely quantifies textual-motion coherence at the clip level. This facilitates direct, fine-grained text-to-motion clip matching and eliminates prevalent redundancy. In the generation stage, a motion prompt module effectively utilizes retrieved motion clips to produce semantically consistent movements. Extensive evaluations confirm that MOST achieves state-of-the-art text-to-motion retrieval and generation performance by comprehensively addressing previous challenges, as demonstrated through quantitative and qualitative results highlighting its effectiveness, especially for rare prompts.
\end{abstract}


\section{Introduction}
Text-to-motion (T2M) generation is a pivotal task in computer animation, which enables synthesis of human-like motion sequences corresponding directly to natural language descriptions, shows potential for creative applications in filmmaking, animation, gaming, VR and robotics \cite{boukhayma2018surface,zhou2025adaptive,kwon2008two,zhou2023hierarchical,wang2019combining,zhou2023multi}. By leveraging advances in deep learning, T2M approaches offer an affordable, data-driven alternative to specialized motion capture. In contrast to restricted motion libraries, generative T2M models enable efficient production of diverse motions tailored to unique prompts. 
Recently, the development of human motion generation methods based on multimodal data has seen significant growth. The multimodal input data include music \cite{kao2020temporally,li2021ai,ren2020self,starke2022deepphase,tseng2023edge,aristidou2022rhythm,fan2011example}, motion categories \cite{guo2020action2motion,petrovich2021action,cervantes2022implicit,guo2022action2video}, motion trajectories \cite{karunratanakul2023guided,wan2023tlcontrol,shafir2023human}, and text \cite{guo2022generating,petrovich2022temos,tevet2023human,chen2023executing,zhang2022motiondiffuse,hou2023two}. Text driven human motion generation has emerged as a recent research focus due to its semantic richness and user-friendliness, which aims to generate motions that are highly consistent with text prompts.

\begin{figure}[t]
    \centering
     \includegraphics[width=\linewidth]{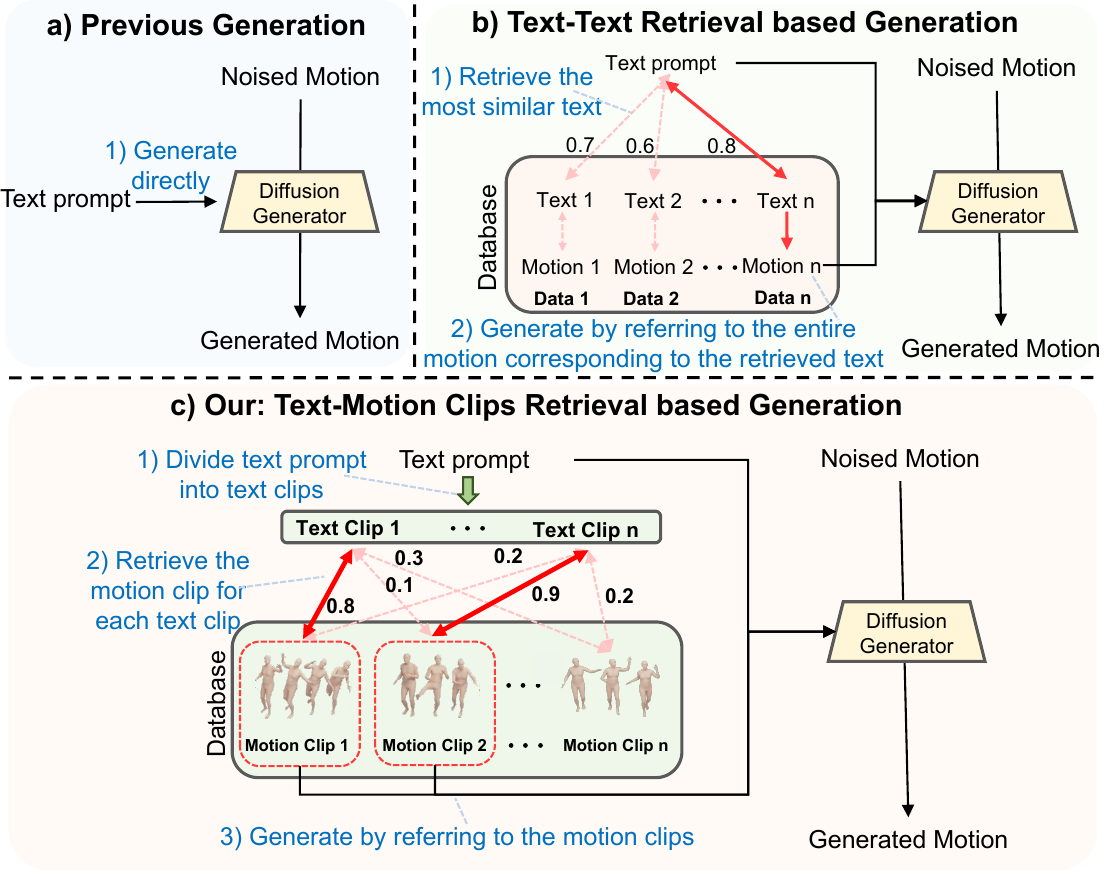}
    \caption{Comparison with existing T2M generation methods. a) Previous methods are limited to simple and common text prompts. b) The motion generated using text-text retrieval suffers from significant similarity between texts, leading to restricted performance. c) In contrast, our approach utilizes text-motion retrieval to leverage common motion clips as prompts for generating high-quality rare motions, effectively guiding the generation process.}
    \label{teaser}
\end{figure}

Researchers have advanced the T2M generation task through various methodologies, including feature alignment between text and motion \cite{lin:vigil18, ahuja2019language2pose,ghosh2021synthesis,petrovich2022temos, bhattacharya2021text2gestures,guo2022generating}, conditional autoregressive models \cite{guo2022tm2t,guo2023momask,zhang2023generating,gong2023tm2d,jiang2024motiongpt}, and diffusion models (Fig. \ref{teaser}a) conditioned solely on text \cite{zhang2022motiondiffuse, chen2023executing, wang2023fg, tevet2022human,kim2023flame}. While these works have addressed numerous challenges, opportunities remain to handle rare text scenarios, specifically involving unseen movements and complex action combinations. Unseen movements represent a motion not encountered during training, while complex action combinations refer to generating a series of multiple different movements. Our research aims to explore techniques for optimizing cross-modal knowledge transfer and multi-source generation to synthesize high-fidelity motion conditioned on rare text descriptions not well addressed by prior methods.

\begin{figure}[t]
    \centering
     \includegraphics[width=\linewidth]{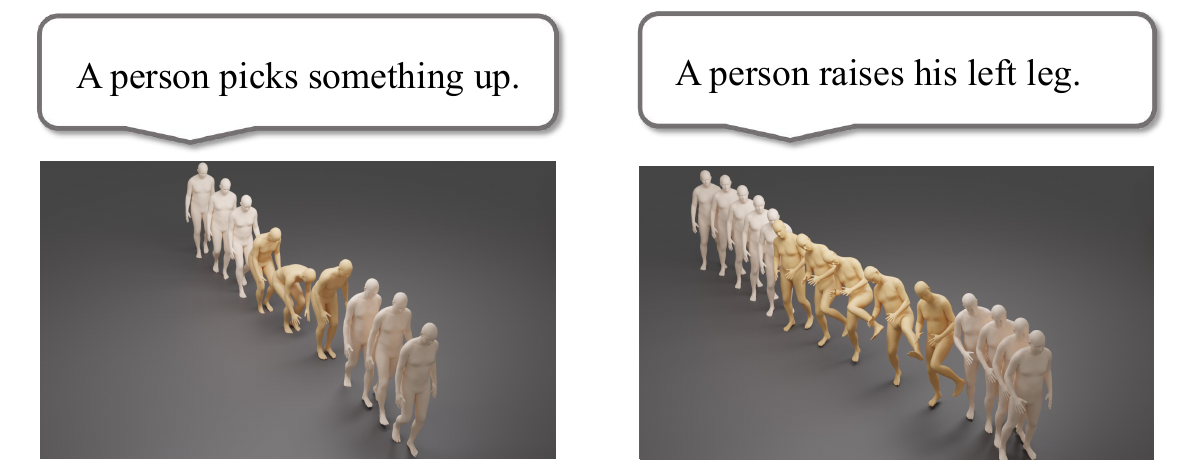}
    \caption{Qualitative Analysis on the Motion Redundancy Problem. We depict the keyframe parts in dark brown and label the remaining parts in light brown.} 
    \label{fig_problem}
\end{figure}

The diffusion-based method Fg-T2M \cite{wang2023fg} utilizes language syntax structure information to construct precise text prompt features for fine-grained human motion generation. GUESS \cite{gao2024guess} employs a cascade-diffusion and gradually enriching synthesis strategy to refine motion details in hierarchical stages. However, they encounter challenges when dealing with rare texts. Techniques like ReMoDiffuse \cite{zhang2023remodiffuse} leverage text-text retrieval (Fig. \ref{teaser}b) to obtain the text most similar to the given text prompt. Using the entire motion sequence corresponding to the retrieval text augments model generalization for rare text. However, enhancing model generalization with the retrieval-based method involves extracting core prior knowledge from motion data to improve the model's generation ability. Therefore, it encounters a key issue: motion redundancy. As shown in Fig. \ref{fig_problem}, the light blue frames beyond the dark blue keyframes lack meaningful information. The essence of one motion often resides within the motion clips refer to short several motion keyframes, whereas the whole motion sequence tends to mask the vital cues. This hinders the model's ability to efficiently learn the retrieve knowledge, ultimately impacting the generation process. Without addressing such issues, current methods may obscure vital cues with redundant motion information, potentially resulting in inaccurate learning and posing challenges with rare text prompts.

We present \textbf{MOST}, a novel two-stage framework addressing key limitations of existing text-to-motion methods, which struggle with coarse-grained matching and redundancy masking important semantic cues. Our insight is that precisely modeling fine-grained clip relationships can overcome these challenges. In the retrieval stage, we introduce the Temporal Clip Banzhaf Interaction formulation, which quantifies textual-motion coherence at the clip level. Drawing from cooperative game theory, this approach treats text and motion clips as ``players,'' computing their marginal contributions through coalition formation. Aligning features of text clips and motion clips by their Banzhaf value at the fine-grained level, we enable direct text-to-clip matching (see Fig. \ref{teaser}c). This method effectively eliminates the redundancy often found in full-sequence retrieval approaches, allowing for more precise retrieval of motion clips. Moreover, we introduce a novel revenue function containing generalization degree ratio to maximize motion information utilization. Treating semantically aligned text-motion clips as interacting "players", our approach yields higher interaction values capturing finer-grained associations beyond the frame-level not achievable by other Banzhaf Interaction methods. 

In \emph{generation stage}, our Motion Prompt Module fuses the retrieved clips with text using the Temporal Banzhaf Interaction-derived clip weights. This provides optimal contextual guidance for our motion transformer to refine noisy inputs. Conditioning on the fused prompts helps address challenges such as generating rare motions described by complex language, while mitigating redundancy versus full sequences. Extensive evaluation on standard datasets HumanML3D~\cite{guo2022generating} and KIT-ML~\cite{plappert2016kit} shows our framework achieves state-of-the-art performance by leveraging fine-grained temporal modeling in the transformer guided by optimal prompts from our two-stage approach. Novel metrics validate MOST resolves issues like redundancy masking semantics in prior work.

Our contributions can be summarized as follows:
\begin{itemize}
  \item A novel framework for text-to-motion generation which incorporates a text-to-motion retrieval stage leveraging rich motion details and a text-to-motion generation stage synthesized by key clips for rare text motions, effectively addressing motion redundancy in existing solutions.
    \item We introduce a temporal clip Banzhaf interaction, grounded in cooperative game theory, to enable fine-grained clip-level text-motion alignment.
  \item We propose a motion diffusion model to effectively employ key motion clips, capture inter-relationships between multiple prompt clips.
  \item Extensive evaluations on HumanML3D and KIT-ML quantitatively and qualitatively validate that MOST achieves superior text-to-motion retrieval and generation, especially for rare prompts.  
\end{itemize}

\section{Related Work}

\subsection{Text-Driven Motion Generation}

Text-driven human motion generation has captured widespread interest in the community, showcasing the vast potential for varied applications. Initial attempts involved learning to map text prompts and motion sequences into a shared multi-modal space. An early standout is Language2Pose \cite{ahuja2019language2pose}, which fused language and motion in the same embedding space. Following this, Ghosh \cite{ghosh2021synthesis} established hierarchical links between language and body, segmenting representations into discrete components for upper and lower body parts. 
To broaden the diversity of motions, some research introduced mechanisms akin to variational autoencoders. TEMOS \cite{petrovich2022temos} merged motion and text variational autoencoders, aligning both through a shared space constrained by Kullback-Leibler divergence loss. Guo \cite{guo2022generating} developed a temporal VAE for iterative motion sequence generation. Nevertheless, a critical limitation of these approaches is forcing the alignment of text and motion into the same spaces, which can dilute information fidelity in both fields.

The emergence of autoregressive models in language and image has sparked innovation in motion generation. TM2T \cite{guo2022tm2t} leverages vector quantized VAEs (VQ-VAE) for mutual mapping between motion and texts. T2M-GPT \cite{zhang2023generating} builds upon this with EMA and code resetting. AttT2M \cite{zhong2023attt2m} introduces a spatio-temporal approach that focuses on body part attention for encoding motion, thereby enhancing motion expressiveness. MMM \cite{pinyoanuntapong2023mmm} transforms human motions into discrete token sequences in latent space and employs a conditional masked motion transformer, which learns to predict masked motion tokens conditioned on texts. Together, these two modules generate high-fidelity human movements. However, the one-way prediction of motion tokens by these models limits their efficiency in capturing bidirectional motion dependencies, resulting in longer training and inference durations.

There are additional methodologies that adopt physics-based strategies. For instance, MoConVQ \cite{yao2024moconvq} utilizes VQ-VAE and model-based reinforcement learning to generate high-quality natural motions. Peng \cite{peng2022ase} introduces adversarial skill embeddings, which learn general and reusable motor skills for physically simulated characters. PhysDiff \cite{yuan2023physdiff} integrates physical constraints into the diffusion process, proposing a novel physics-guided motion diffusion model that generates realistic motions without artifacts. However, these methods depend on the manual design of complex reward functions and involve a lengthy training process.

Existing methods \cite{guo2022generating,guo2022tm2t} primarily utilize implicit alignment between language latents and motion latents. However, some approaches explicitly align these latents to enhance the subsequent motion generation process. For instance, PoseScript \cite{delmas2024posescript} extracts low-level posture information through a set of simple yet versatile 3D keypoint rules, which are then combined into higher-level text descriptions using grammatical rules. CoMo \cite{huang2024controllable} decomposes motion into discrete, semantically meaningful pose codes, with each code representing the state of specific body parts at a given moment. This explicit representation allows for intuitive interaction and modification of sequences to generate compliant motion. However, such explicit alignment methods are often constrained to local motion or trained datasets, which can limit their generalizability to rare text prompts.

Recent diffusion models like \cite{zhang2022motiondiffuse, wang2023fg, tevet2022human,kim2023flame,gao2024guess,wang2025fg} adopt conditional diffusion models to learn the probabilistic mapping from text prompts to motion sequences. MLD \cite{chen2023executing} draws inspiration from latent diffusion models \cite{rombach2022high} and further proposes the motion latent diffusion model. 
GUESS \cite{gao2024guess} employs a cascading diffusion framework and gradually enriching synthesis strategy, recursively abstracting human poses and refining motion details in phases.
Moreover, by incorporating a dynamic multi-condition fusion mechanism, it enhances the interaction between text prompts and synthesized motions, optimizing the cross-modal motion synthesis task. Current methods encounter difficulties with rare texts. Despite ReMoDiffuse's utilization of retrieval technology \cite{zhang2023remodiffuse} to augment generation with additional knowledge from retrieved samples, motion redundancy within entire sequences hinders generalization for rare prompts. Innovations are necessary to enhance representation of complex cross-modal associations, particularly for fine-grained uncommon text prompts, and address existing limitations to improve modeling capabilities in challenging domains.

\subsection{Text-to-Motion Retrieval}

The text-to-motion retrieval task has seen notable advances. MoT \cite{messina2023text} pioneered joint text-motion encoding using CLIP \cite{radford2021learning} and transformers \cite{vaswani2017attention}, employing partitioned spatiotemporal attention to effectively aggregate different skeletal joints in space and time. TMR \cite{petrovich23tmr} extends text-to-motion generation model TEMOS \cite{petrovich2022temos}, incorporating contrastive loss while maintaining motion generation loss to better construct cross modal latent spaces. While progressing the field, opportunities remain to systematically represent fine-grained associations essential for precise semantic alignment, important for rare text diffusion modeling. The current methods still remain at the entity-level, which explores the relationships between the overall texts and the whole motion sequences. However, solely considering coarse entity-level semantics overlooked detailed clip-level correspondences within pairs.  Addressing oversimplified entity-level matching by incorporating clip intersections at finer scales could strengthen modeling for rare text prompts.

\subsection{Cooperative Game Theory in Vision-Language}

Cooperative game theory provides technical frameworks for modeling collaborative multi-element systems. Prior works incorporated interaction values with differing outcomes in vision-language domains. LOUPE \cite{li2022fine} utilizes Shapley interaction \cite{sun2020random} and uncertainty-aware neural Shapley interaction learning module for visual-language pre-training framework, which is the first method to learn semantic alignment from a new perspective of game theory. TG-VQA \cite{li2023tg} employed Banzhaf interaction \cite{marichal2011weighted} for video, question, and answer as ternary players, to  simulate complex relationships between multiple players for VideoQA task. HBI \cite{jin2023video} also used Banzhaf interaction \cite{marichal2011weighted} for video-text representation to value possible correspondence between video frames and text words for sensitive and explainable cross-modal contrast.
However, existing measures fail to guide rare text motion diffusion in T2M retrieval under coarse-grained entity-level, necessitating fine-grained clip-level relationships. Our temporal clip Banzhaf interaction builds on quantifying coalitions by marginal exclusion assessment, overcoming limitations of entity-level formulations. By rigorously evaluating detailed text-motion clip interdependencies, our approach facilitates comprehensive knowledge modeling, critical to multimedia element retrieval and alignment objectives. Leveraging the theoretical foundations of our refined formulation addresses prior art shortcomings.

\section{Preliminaries}

\subsection{Cooperative Game Theory}
In cooperative game theory, a group of players $N={1,2,...,n}$ collaborates to maximize revenue $v(.)$ \cite{grabisch1999axiomatic}. In T2M tasks, $v$ quantifies the similarity between elements such as text clips and motion clips. These players can represent elements extracted from texts and motions. Previous works have focused on aligning entire text-motion entities. However, for rare texts describing complex motions, establishing fine-grained cooperative relationships between elements becomes crucial.

\textbf{Banzhaf Values.} This metric provides insights into the roles of players in cooperative games.
The Banzhaf value assesses each player's average additional impact across different player permutations. Specifically, player $i$'s Banzhaf value $B(i|N)$ is computed across coalitions $S \subseteq N$ without player $i$:
\begin{equation}
B(i|N) = \sum_{S \subseteq N \backslash \{i\}} \rho(S)(v(S \cup \{i\}) - v(S))
\end{equation}
Here, $\rho(S) = \frac{1}{2^{n-1}}$ denotes coalition likelihood, and $N \backslash {i}$ represents $N$ excluding player $i$. This formulation systematically evaluates player contributions in diverse cooperative contexts.

\textbf{Banzhaf Interaction.} In a cooperative game, some players tend to form a coalition. Banzhaf interaction quantifies the incremental contribution of coalitions \cite{grabisch1999axiomatic}. In our method, text clips and motion clips are represented as "players". Banzhaf interactions between clip coalitions are optimized to precisely align informative elements described in rare texts. 
By framing T2M as a cooperative game and utilizing Banzhaf interaction to capture complex collaborative patterns, we overcome previous methods' limitations in handling rare text descriptions with an interpretable framework.

For T2M generation tasks, Banzhaf interaction can be utilized to model relationships between players which could be either text clips or motion clips. For a coalition $\{i,j\}$ involving any two players $i$ and $j$ where $i,j \in N$ and $N$ denotes the set of all players which includes both text clips and motion clips, we treat $\{i,j\}$ as a unified player representing their joint participation. Then, we remove the individual players $i$ and $j$ from the game and add the coalition $\{i,j\}$ to the game. The Banzhaf interaction $I(\{i,j\})$ for coalition $\{i,j\} \subseteq N$ is:
\begin{equation}
\begin{aligned}
    I(\{i,j\}) = \sum_{S \subseteq N \backslash \{i,j\}} \rho(S) [(v(S \cup \{i,j\}) + v(S) \\
    - v(S \cup \{i\}) - v(S \cup \{j\})]
\end{aligned}
\label{banzhaf_inter}
\end{equation}
where $\rho(S) = \frac{1}{2^{n-2}}$ is the likelihood of coalition $S$ being sampled, and $N \backslash \{i,j\}$ denotes the set obtained by removing players $i$ and $j$ from $N$. 
$I(\{i,j\})$ depicts the interaction between two players $i$ and $j$ in coalition $\{i,j\}$. A higher $I(\{i,j\})$ indicates closer cooperation, while a lower value suggests weaker cooperation. This formulation models text-motion relationships, aiding T2M generation.
We propose using the Banzhaf interaction to capture fine-grained alignment between text and motion clips in T2M Retrieval. This enhances retrieval and boosts model performance with rare text inputs.

\subsection{Diffusion Model}
Diffusion models \cite{ho2020denoising} form the foundation of our research on text-to-motion generation tasks. It embraces a Markovian framework comprising of a forward noise process and a reverse denoising process. The forward process starts with the initial real motion data $x_0$ at step 0, and progressively adds Gaussian noise at each time step $t$, represented as:
\begin{equation}
q(\mathbf{x}_t \vert \mathbf{x}_{t-1}) = \mathcal{N}(\mathbf{x}_t; \sqrt{1-\beta_t}\mathbf{x}_{t-1}, \beta_t I)
\end{equation}
where $\beta_t$ is a hyperparameter, effectively converting the original movement sequence into a distribution approximating $\mathcal{N}(0, I)$. The entire diffusion process is formulated as:
\begin{equation}
        q(\mathbf{x}_{1:T} \vert \mathbf{x}_0) \,=\, \prod_{t=1}^{T} q(\mathbf{x}_t \vert \mathbf{x}_{t-1})
\end{equation}
where $T$ denotes the total steps in diffusion. The reverse process aims to eliminate this added noise from $x_t$ to revert back to $x_0$, defined as: 
\begin{equation}
p_{\theta}(\mathbf{x}_{0:T} \vert c,r) = p(\mathbf{x}_{T} \vert c,r) \prod_{t=1}^{T}p_{\theta}(\mathbf{x}_{t-1} \vert \mathbf{x}_{t},c,r),
\end{equation}
\begin{equation}
p_{\theta}(\mathbf{x}_{t-1} \vert \mathbf{x}_{t},c,r) = \mathcal{N}(\mathbf{x}_{t-1}; \mu_\theta(\mathbf{x}_{t}, t,c,r), \Sigma_\theta(\mathbf{x}_{t}, t,c,r)).
\end{equation}
in our retrieval-based text-to-motion generation, $c$ denotes the given text prompts and $r$ represents the retrieved motion references. By incorporating the reverse denoising process guided by Text-to-Motion Retrieval, we effectively obtain key motion clips, which serve as motion prompts in Text-to-Motion Generation, enabling motion generation under rare text conditions. This novel approach significantly enhances the capabilities of our proposed MOST framework and outperforms existing methods on T2M retrieval and generation tasks by effectively addressing motion redundancy challenges.

\begin{figure}[t]
    \centering
     \includegraphics[width=\linewidth]{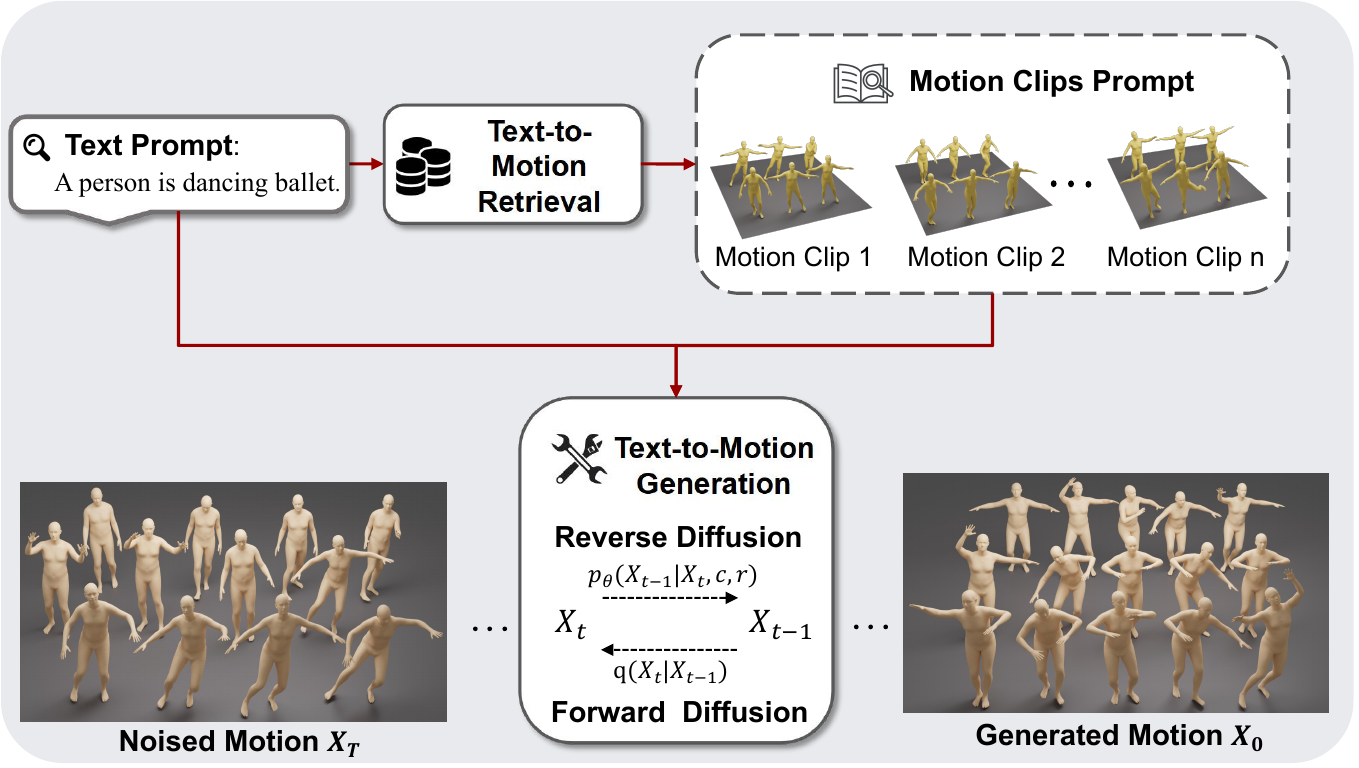}
    \caption{\textbf{Overview of our MOST.} This figure illustrates the reverse denoising process of the motion diffusion model. The first step is using Text-to-Motion Retrieval to obtain motion clips. These clips along with motion and text prompts are then leveraged in the Text-to-Motion Generation stage to generate motion.  
    }
    \label{pipeline}
\end{figure}

\begin{figure*}[t]
    \centering
     \includegraphics[width=\linewidth]{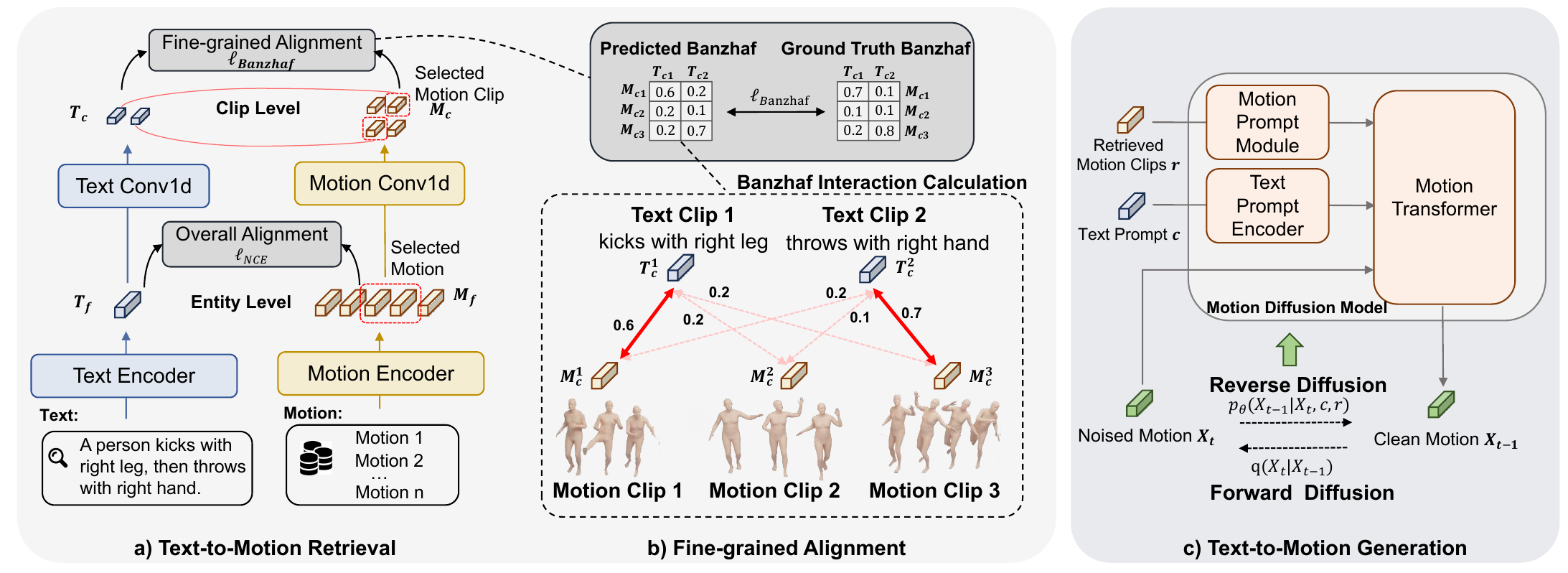}
    \caption{ a) \textbf{Text-to-Motion Retrieval.} In this stage, a dual stream encoder and $\mathcal{L}_\text{NCE}$ are employed to achieve entity alignment. Additionally, using a conv1d layer to obtain text-motion features at the clip level. Fine-grained alignment of each text-motion clip is performed using the $\mathcal{L}_\text{Banzhaf}$ loss function. b) \textbf{Fine-grained Alignment.} All text clips and motion clips act as players, calculating the temporal clip Banzhaf values between each text clip and each motion clip through the defined revenue value function. Constraining the predicted values against the ground truth to achieve fine-grained alignment of text and motion. c) \textbf{Text-to-Motion Generation}. This stage involves the motion prompt module, which integrates and comprehends the retrieved motion clips. The motion features $X_t$, text features $c$, and prompt clip features $r$ are fed into the motion transformer decoder to generate the cleaned motion sequence $X_{t-1}$.
    }
    \label{pipeline-detail}
\end{figure*}

\section{Proposed Method: MOST}

We address the challenge of generating realistic motions for rare text prompts, where prior methods struggle due to motion redundancy. Our MOST tackles this issue through a two-stage approach (Fig. \ref{pipeline}). In the \emph{retrieval stage}, we leverage the temporal clip Banzhaf interaction $I(\{i,j\})$ as defined by Equation \ref{banzhaf_inter}, enabling fine-grained clip-level alignments between text and motion. This allows MOST to directly capture motion clips containing vital information, effectively resolving motion redundancy. The \emph{generation stage} introduces a motion prompt module weighted by $I(\{i,j\})$ to guide our forward-T reverse-0 diffusion process, restoring the original motion $X_0$. This module effectively aggregates and utilizes key motion clips, further addressing motion redundancy. Extensive quantitative retrieval and generation results validate that MOST surpasses the state-of-the-art (SOTA) by addressing motion redundancy through our two-stage formulation, leveraging finer clip-level motion structure.

\subsection{Text-to-Motion Retrieval}

Our method introduces a novel retrieval stage in Fig. \ref{pipeline-detail}a, aimed at directly retrieving key motion clips from the input text, thus mitigating limitations observed in prior works plagued by motion redundancy. A significant advantage of this approach lies in its capacity to merge frames into learnable clips, thereby preserving continuous temporal flow, rather than accumulating redundancy as seen in methods like KNN clustering \cite{jin2023video}, which treat frames as static snapshots. By focusing on modeling temporal information at the clip-level, our approach effectively addresses challenges encountered by other Banzhaf Interaction techniques that operate solely on individual frames. We will now explore the dual alignment processes, Banzhaf formulation, and motion clip selection to elaborate our approach's contributions.

\subsubsection{Text-Motion Overall Alignment}
Effective motion retrieval in T2M tasks requires a comprehensive approach to align multi-modal text and motion representations, supervised by contrastive loss. During feature extraction, a text encoder maps input text into features $T_f=\{t^j\}_{j=1}^{T_w}$ capturing textual semantics, while a motion encoder outputs features $M_f=\{m^i\}_{i=1}^{T_m}$ encoding temporal dynamics, with $T_w$ or $T_m$ denoting the length of text or motion. Noise contrastive estimation (NCE) loss $\mathcal{L}_{NCE}$ \cite{oord2018representation} is employed for contrastive learning, maximizing the similarity between paired representations measured by similarity score $S_{m,t}$ while minimizing non-paired similarity.
Precisely, it is defined as: 
\begin{equation}
\begin{aligned}
\mathcal{L}_\text{NCE} = - \frac{1}{2B_s}[\sum_{k=1}^{B_s} \log  \frac{\exp(S_{m_k,t_k} / \tau)} {\sum_{l=1}^{B_s} \exp(S_{m_k,t_l} / \tau) } \\
+ \log \frac{\exp(S_{t_k,m_k} / \tau)} {\sum_{l=1}^{B_s} \exp(S_{t_k,m_l} / \tau)} ]
\end{aligned}
\end{equation}
where $B$ is the batch size and $\tau$ is the temperature hyper-parameter. Besides, $S_{m,t}$ is defined as:
\begin{equation}
S_{m,t} =  \frac{1}{2} \left(\frac{1}{T_m} \sum_{i=1}^{T_m} \max_j A_{ij} + \frac{1}{T_w} \sum_{j=1}^{T_w} \max_i A_{ij}\right)
\end{equation}
where $A_{ij} = (m^{i})^\mathrm{T} (t^{j})$.
This oversees entity-level matching, but existing methods overlook the detailed alignment required for our task. We hence introduce dual alignment with $\mathcal{L}_{NCE}$ and our clip-level $\mathcal{L}_B$ loss to facilitate feature matching for precise clip retrieval, guiding motion generation.

\subsubsection{Text-Motion Fine-Grained Alignment}

Entity-level alignment alone falls short in our text-to-motion retrieval task, where identifying pivotal motion clips corresponding to the input text is essential. To address this, we introduce fine-grained alignment at the temporal clip level. In creating these temporal clips, our aim is to combine consecutive motion frames into higher-level, semantics-preserving clips for a comprehensive motion representation. Given the sequential nature of motions, we utilize a 1D convolutional (conv1d) layer followed by self-attention to convert the motion features $M_f=\{m^i\}^{T_m}_{i=1}$ into lower-resolution clips $M_c=\{m^i\}^{T_s}_{i=1}$, with $T_s$ representing the length of clips and $T_s<T_m$. Similarly, text features $T_f=\{t^j\}^{T_w}_{j=1}$ are transformed into text clips $T_c=\{t^j\}^{T_s}_{j=1}$. 

To facilitate fine-grained clip interaction, we introduce the Temporal Clip Banzhaf Interaction, laying the theoretical groundwork for our approach. This interaction constrains clip representations to maximize similarity between correlated text-motion clips during retrieval, enabling both entity-level matching via $\mathcal{L}_{\text{NCE}}$ and clip-level matching. Comprising three key components as follows, this interaction facilitates the precise identification of key motion clips guided by the input text.

\textbf{1. Player Setting:}
To model clip cooperation, we define clips as players:
\begin{equation}
N=\{m^i\}^{T_s}_{i=1} \cup \{t^j\}^{T_s}_{j=1}
\end{equation}
As embedded representations of segmented semantics, clips exhibit collaborative relationships during retrieval. Representing them as player set allows quantifying these interactions.

\textbf{2. Revenue Value Function:}
This payoff function indicates the benefits derived from collaboration, with a higher value signifying stronger correspondence and encouraging closer cooperation for increased payoffs during retrieval. Conversely, uncorrelated clips do not benefit from collaboration. We base this function on the similarity score $S_{m,t}$, which measures the semantic alignment between the embedded contents of two clips. To enhance motion richness, we propose calculating the generalization degree ratio for both motion and text clips:
\begin{equation}
\text{W} = \text{W}_m (\text{W}_w)^{\mathsf{T}},
\end{equation}
\begin{equation}
\text{W}_w =\frac{1}{l_w} \sum_{j=2}^{l_w} \lvert t^{j} - t^{j-1}   \rvert, \; \; \text{W}_m  =\frac{1}{l_m} \sum_{j=2}^{l_m} \lvert m^{j} - m^{j-1} \rvert
\end{equation}
where $l_m$ = $l_w$, $l_m$and $l_w$ are the numbers of frames within a clip. A lower generalization degree ratio indicates less information content and higher redundancy within a clip, and vice versa. Thus, we obtain the revenue value function: 
\begin{equation}
v =  \frac{1}{2} \left(\frac{1}{T_m} \sum_{i=1}^{T_m} \max_j \text{W} A_{ij} + \frac{1}{T_w} \sum_{j=1}^{T_w} \max_i \text{W} A_{ij}\right)
\end{equation}
which considers enhancing motion richness while reducing the proportion of motion redundancy.

\textbf{3. Temporal Clip Banzhaf Computation:} 
Our objective is to compute the true Banzhaf interaction value $I(\{m, t\})$ for each clip pair coalition $\{m, t\}$ using the Banzhaf formula. To estimate these values accurately, we develop a predictor P for temporal clip Banzhaf interaction and align it with the ground truth through the following loss:
\begin{equation}
\mathcal{L}_\text{B} = - \sum_{i=1}^{T_s} (I_{t2m}^{i}) \log (P_{t2m}^{i}) - \sum_{i=1}^{T_s} (I_{m2t}^{i}) \log (P_{m2t}^{i})
\end{equation}

This cross-entropy loss optimizes the distributions between ground truth $I$ and predicted $P$, training the predictor to infer interaction strengths accurately. $\mathcal{L}_\text{B}$ ensures precise alignment between predicted and actual Banzhaf scores, establishing fine-grained semantic correlations. During inference, the predictor computes interaction values to select the motion clip with the highest $P$ for each text clip. This dual-level matching via predicted Banzhaf interactions addresses existing problems by enhancing retrieval through precisely identifying the most relevant motion content.

\subsubsection{Selection of Key Motion Clips}

To mitigate motion redundancy in retrieval, we select key motion clips over full sequences for generation. Initially, we identify the top $K_e$ motion entities using the similarity function $S_{m,t}$, followed by partitioning text/entities into $T_s$ clips utilizing conv1d. While computing the text-motion alignment matrix $A$ establishes clip relationships, existing methods often struggle to precisely match rare text clips. To address this, we introduce a temporal clip Banzhaf interaction predictor, modeling cooperative clip patterns via cooperative game theory. This formulation is pivotal for understanding complex rare text descriptions, calculating interaction values that quantify each clip coalition's contribution. Motion clips corresponding to each text clip are then sorted by interaction scores, selecting the top $n$ most cooperative clip ids with maximum semantic guidance. Then extracting corresponding motion clips from the original motion sequence based on the number id and clip length $T_s$. For example, if text clip 1 corresponds to the motion clip with the highest similarity, which is id 0, and $T_s$ is 39, then the motion from 0*39 frame to 1*39 frame of the original motion sequence will be extracted. Compared to full sequences, these key motion clips utilize preserved semantic knowledge more effectively, enabling better motion generation through conditioned generation.

\subsubsection{Training Objective}

We propose a combined training objective comprising cross-modal NCE loss $\mathcal{L}_\text{NCE}$ and temporal clip Banzhaf interaction loss $\mathcal{L}_\text{B}$, with $\mathcal{L}_\text{B}$ addressing the incremental contributions of clip coalitions through Banzhaf interaction values, crucial for precise fine-grained clip alignments with rare texts. This overall retrieval loss:
\begin{equation}
\mathcal{L}_\text{R} = \mathcal{L}_\text{NCE} + \lambda_\text{B} \mathcal{L}_\text{B}
\end{equation}
ensures the capture of both overall and clip-level cross-modal alignments, facilitating an enhanced retrieval process by leveraging rich motion details. The weighting hyperparameter $\lambda_\text{B}$ optimizes this combination of losses to effectively address the research problem of text-to-motion matching and contribute to improved retrieval performance, particularly for rare texts.

\subsection{Text-to-Motion Generation via Selective Clip Prompts}

\subsubsection{Motion Generation via Diffusion Models}
Conventional generation approach \cite{zhang2023remodiffuse} that relies on full motion sequences as prompts often introduce motion redundancy, masking critical cues. To address this, we leverage compact yet informative key clips retrieved through Banzhaf interaction modeling. As illustrated in Fig. \ref{pipeline-detail}b, our motion diffusion model acts as the main denoising model within the total diffusion process, aiming to transform noisy $X_t$ sequences into denoised $X_{t-1}$ at each stage. Comprising a motion prompt module, text prompt encoder, and motion transformer, our motion diffusion model integrates clip coalitions with text, preserving representative motion information while mitigating redundancy. Empowered by this enhanced contextual guidance, our generator overcomes challenges posed by prior works. Moreover, the text encoder accurately and efficiently extracts rich rare text features, while the motion transformer improves the quality of motions conditioned on rare texts, capturing semantics more effectively through the utilization of correlated motion knowledge. By incorporating only the most pivotal clips to augment inputs, this two-stage approach leverages selectively preserved details to enhance generation.

MOST is learned with classifier-free diffusion guidance \cite{ho2022classifier} which scales conditional and unconditional distributions with hyperparameter $s$:

\begin{equation}
\epsilon = s\epsilon_\theta(x_t,t,c,r)+(1-s)\epsilon_\theta(x_t,t,\varnothing)
\end{equation}
where $t$, $c$, and $r$ represent timesteps, text prompts, and retrieved motion clips, respectively.
By optimizing representations and leveraging text/clip semantics, the motion transformer effectively overcomes limitations inherent in refinement processes found in previous works.

To optimize our motion diffusion model, we aim to predict the clean state $X_0$. The training process minimizes:
\begin{equation}
       \mathcal{L}=\mathrm{E}_{t \in [1,T], \mathbf{x}_0 \sim q(\mathbf{x}_0),\epsilon \sim \mathcal{N}(0, I)} [\parallel \mathbf{x}_0 - \epsilon_\theta(\mathrm{x}_t,\mathrm{t},\mathrm{c},\mathrm{r}) \parallel_2^2]
\end{equation}
which involves minimizing the difference between the predicted and ground truth motions. This approach enables the model to learn to generate high-quality denoised motion sequences.

We now explore the motion prompt module, the text prompt encoder, and the architecture of the motion transformer, which collectively synthesize motions from the prompts.

\subsubsection{Motion Prompt Module}

This module is introduced to overcomes limitations observed in previous methods \cite{zhang2023remodiffuse}, which often neglected the inter-relationships between multiple prompt entities, resulting in a lack of coherent motion context. Our module fuses motion knowledge by leveraging retrieved clips as additional contextual clues, thereby minimizing motion redundancy and enriching understanding. For each text clip, we retrieve the top-$K_c$ motion clips and calculate the temporal Banzhaf interaction between each clip and text to obtain weights via softmax. The prompt motion clip $R_i$ for the $i^{th}$ text clip is then derived as the weighted summation of the top clips using our theoretical formulation:
\begin{equation}
R_i = \sum_{m=1}^{K_c} \frac{\exp(B_{m})r_m}{\sum_{n=1}^{K_c}\exp(B_{n})}
\end{equation}
where $r_m$ is a retrieved clip, and $B_{m}$ is its Banzhaf interaction with the text clip.
This fusion approach effectively preserves semantics from multiple related clips, empowering guided generation to overcome challenges associated with motion redundancy and lack of contextual richness. Consequently, our approach contributes to enhanced generation, enabling better matching of complex descriptions by leveraging retained motion knowledge.

\subsubsection{Text Prompt Encoder}
We construct our text encoder similarly to MotionDiffuse \cite{zhang2022motiondiffuse}, based on transformer layers. Initially, we use parameter weights from CLIP \cite{radford2021learning} to configure the first few layers of the text encoder, subsequently freezing these layers to leverage their robust generalization and encoding capabilities. Additionally, we integrate classical transformer layers \cite{vaswani2017attention} to further enhance and enrich the textual features.

\subsubsection{Motion Transformer}

The Motion Transformer incorporates text/motion prompts and inputs across attention-based layers. Each contains:

\begin{enumerate}
\item Self-attention: Standard implementation \cite{shen2021efficient}.

\item Cross-attention with text prompts \cite{shen2021efficient}. 

\item Cross-attention with motion prompts: A novel approach. We partition $X_t$ into $T_S$ clips $X_t^{i}$. Each clip's noisy features serve as queries to attend over the corresponding prompt motion features $R_i$ as keys/values.
\end{enumerate}

This formulation discerns and leverages valuable prompt knowledge from multiple clips, enhancing denoising, which is a key challenge.

\section{Experiments}

\subsection{Experimental Settings}
\subsubsection{Datasets}
Given the specific tasks we address in text-to-motion retrieval and generation, we utilize both the KIT-ML \cite{plappert2016kit} and HumanML3D datasets \cite{guo2022generating} for our experiments. The HumanML3D dataset combines data from HumanAct12 \cite{guo2020action2motion} and AMASS \cite{mahmood2019amass}, providing three text descriptions for each motion, averaging 12 words per description. Encompassing various activities such as daily tasks, sports, and acrobatics, it comprises 14,616 motions and 44,970 texts, totaling approximately 28.59 hours. In contrast, the KIT Motion-Language (KIT-ML) dataset \cite{plappert2016kit} serves as a smaller-scale benchmark, featuring motions described by one to four sentences, averaging 8 words per sentence. This dataset includes 3,911 motions and 6,353 descriptions, amounting to about 10.33 hours.

\subsubsection{Metrics}
In the T2M retrieval task, our evaluation metrics include recall at various ranks (R@k) and Median Rank (MedR). R@k indicates the percentage of times the correct label appears among the top $K$ results, while Median Rank (MedR) denotes the median position of the correct label in the ranked results.

For the T2M generation, we employ the following metrics:
(1) R-TOP assesses the similarity between the generated motion and the text prompt by calculating the probability of the true prompt being ranked within the top $K$.
(2) FID measures the likeness between the feature distributions extracted from the generated motions and the ground truth.
(3) MM Dist computes the average Euclidean distance between the feature of generated motions and the text prompt feature.
(4) Diversity evaluates the dissimilarity among all generated motions across all descriptions.
(5) Multi-Modality measures the average variance of generated motions for a given text prompt.
\begin{table}[t]
    \centering
\large
    \resizebox{\linewidth}{!}{
    \begin{tabular}{cl|cccccc}
        \toprule
         \multirow{2}{*}{\textbf{Protocol}} & \multirow{2}{*}{\textbf{Methods}} & \multicolumn{6}{c}{\textbf{Text-motion retrieval}}  \\
         & & \small{R@1 $\uparrow$} & \small{R@2 $\uparrow$} & \small{R@3 $\uparrow$} &  \small{R@5 $\uparrow$} & \small{R@10 $\uparrow$} & \small{MedR $\downarrow$}  \\
\midrule 
\multirow{4}{*}{HumanML3D } & TEMOS\cite{petrovich2022temos} &  2.12 &  4.09 &  5.87 &  8.26 & 13.52 & 173.0 \\
&        Guo et al.\cite{guo2022generating}&  1.80 &  3.42 &  4.79 &  7.12 & 12.47 & 81.00  \\
\multirow{2}{*}{(All) }&    TMR\cite{petrovich23tmr} &  5.68 & 10.59 & 14.04 &    \textbf{20.34} & 30.94 & 28.00      \\

& MOST (w/o Banzhaf)&  5.59 &  9.13 &  12.33 &  18.01 & 28.92 &29.00\\
              &    MOST &         \textbf{6.61} &    \textbf{10.82} &    \textbf{14.53} &19.97 &   \textbf{32.05}&    \textbf{25.00}  \\

\midrule
\multirow{3}{*}{HumanML3D }    &TEMOS\cite{petrovich2022temos} & 40.49 & 53.52 & 61.14 & 70.96 & 84.15 &2.33 \\
&       Guo et al.\cite{guo2022generating} & 52.48 & 71.05 & 80.65 & 89.66 & 96.58&1.39  \\
\multirow{1}{*}{(Small Batches) }&     TMR\cite{petrovich23tmr}& 67.16& 81.32 & 86.81 & 91.43 & 95.36 &  1.04 \\
&  MOST &    \textbf{69.20} &   \textbf{84.59} &   \textbf{90.50} &   \textbf{95.26} &   \textbf{98.53} &   \textbf{1.01}  \\

\midrule
\multirow{3}{*}{KIT-ML } & TEMOS\cite{petrovich2022temos}&  7.11 & 13.25 & 17.59 & 24.10 & 35.66 & 24.00  \\
             &     Guo et al.\cite{guo2022generating}&  3.37 &  6.99 & 10.84 & 16.87 & 27.71 & 28.00  \\
\multirow{1}{*}{(All) }&                      TMR\cite{petrovich23tmr} &  7.23 & 13.98 & 20.36 & 28.31 & 40.12 & 17.00  \\
              &   MOST &     \textbf{9.87} &   \textbf{16.31} &   \textbf{23.53} &   \textbf{31.16} &   \textbf{44.11} &   \textbf{11.00}  \\

\midrule
\multirow{3}{*}{KIT-ML }      & TEMOS\cite{petrovich2022temos} & 43.88 & 58.25 & 67.00 & 74.00 & 84.75 &  2.06  \\
&      Guo et al.\cite{guo2022generating}& 42.25 & 62.62 & 75.12 & 87.50 & 96.12 &  1.88  \\
\multirow{1}{*}{(Small Batches) }&           TMR\cite{petrovich23tmr} & 49.25 & 69.75 & 78.25 & 87.88 & 95.00 &  1.50  \\
& MOST &     \textbf{56.49} &   \textbf{76.90} &   \textbf{86.07} &   \textbf{93.41} &   \textbf{97.64} &   \textbf{1.23} \\

    \bottomrule        
    \end{tabular}
    }
    \caption{Text-to-motion retrieval results on both the KIT-ML \cite{plappert2016kit} and HumanML3D \cite{guo2022generating} datasets. Best results are bolded.}
    \label{retrieval_h3d_kit}
\end{table}

\begin{table*}[t]

\centering
\resizebox{\linewidth}{!}{
\begin{tabular}{clcccccccc}
\toprule
\multirow{2}{*}{Datasets} & \multirow{2}{*}{Methods} &\multirow{2}{2cm}{\centering Publication}& \multirow{2}{*}{R-TOP1  $\uparrow$} & \multirow{2}{*}{R-TOP2  $\uparrow$} & \multirow{2}{*}{R-TOP3  $\uparrow$}                                                                &\multirow{2}{*}{FID$\downarrow$} & \multirow{2}{*}{MM Dist$\downarrow$}              & \multirow{2}{*}{Diversity$\rightarrow$}           & \multirow{2}{*}{MultiModality$\uparrow$}           \\
                            \\ \midrule
&Real &
    - &
  $0.511^{\pm.003}$ &
  $0.703^{\pm.003}$ &
  $0.797^{\pm.002}$ &
  $0.002^{\pm.000}$ &
  $2.974^{\pm.008}$ &
  $9.503^{\pm.065}$ &
  \multicolumn{1}{c}{-}
  \\

\multirow{15}{*}{HumanML3D}&
TEMOS \cite{petrovich2022temos}&
    ECCV 2022 &
  $0.424^{\pm.002}$ &
  $0.612^{\pm.002}$ &
  $0.722^{\pm.002}$ &
  $3.734^{\pm.028}$ &
  $3.703^{\pm.008}$ &
  $8.973^{\pm.071}$ &
  $0.368^{\pm.018}$ \\
&
  Temporal VAE \cite{guo2022generating}&
      CVPR 2022 &
    $0.455^{\pm.003}$ &
  $0.636^{\pm.003}$ &
  $0.740^{\pm.003}$ &
  $1.067^{\pm.002}$ &
  $3.340^{\pm.008}$ &
  $9.188^{\pm.002}$ &
  $2.090^{\pm.083}$ \\
&
  MLD \cite{chen2023executing}&
    CVPR 2023 &
  $0.481^{\pm.003}$ &
  $0.673^{\pm.003}$ &
  $0.772^{\pm.002}$ &
  $0.473^{\pm.013}$ &
  $3.196^{\pm.010}$ &
  ${9.724}^{\pm.082}$ &
  $2.413^{\pm.079}$ \\
     &T2M-GPT \cite{zhang2023generating}&
       CVPR 2023 &
    $0.491^{\pm.003}$ &
  $0.680^{\pm.003}$ &
   $0.775^{\pm.002}$ &
    ${0.116}^{\pm.004}$ &
     $3.118^{\pm.011}$ &
      ${9.761}^{\pm.081}$ &
       $1.856^{\pm.011}$ \\
       &MDM \cite{tevet2022human}&
    ICLR 2023 &
  $0.320^{\pm.005}$ &
  $0.498^{\pm.004}$ &
  $0.611^{\pm.007}$ &
  $0.544^{\pm.044}$ &
  $5.566^{\pm.027}$ &
  $9.559^{\pm.086}$ &
  $\underline{2.799}^{\pm.072}$ \\
    &Fg-T2M \cite{wang2023fg}&
        ICCV 2023 &
    $0.492^{\pm.002}$ &
  $0.683^{\pm.003}$ &
   $0.783^{\pm.002}$ &
    $0.243^{\pm.019}$ &
     $3.109^{\pm.007}$ &
      $9.278^{\pm.072}$ &
       $1.614^{\pm.049}$ \\
            &ReMoDiffuse \cite{zhang2023remodiffuse}&
                ICCV 2023 &
    ${0.510}^{\pm.005}$ &
  ${0.698}^{\pm.006}$ &
   ${0.795}^{\pm.004}$ &
    ${0.103}^{\pm.004}$ &
     ${2.974}^{\pm.016}$ &
      $9.018^{\pm.075}$ &
       $1.795^{\pm.043}$ \\

  &MotionGPT \cite{jiang2024motiongpt}&
      NeurIPS 2024 &
  ${0.492}^{\pm.003}$ &
  $0.681^{\pm.003}$ &
  $0.778^{\pm.002}$ &
  $0.232^{\pm.008}$ &
  ${3.096}^{\pm.008}$ &
  $\textbf{9.528}^{\pm.071}$ &
  $2.008^{\pm.084}$ \\
  &FineMoGen \cite{zhang2024finemogen}&
      NeurIPS 2024 &
$0.504^{\pm.002}$ &
$0.690^{\pm.002}$ &
$0.784^{\pm.002}$ &
$0.151^{\pm.008}$ &
$2.998^{\pm.008}$ &
$9.263^{\pm.094}$ &
$2.696^{\pm.079}$ \\
  &AvatarGPT \cite{zhou2023avatargpt}&
      CVPR 2024 &
$0.510^{\pm.005}$ &
$0.702^{\pm.005}$ &
$0.796^{\pm.003}$ &
$0.168^{\pm.008}$ &
- &
$9.624^{\pm.054}$ &
- \\
  &MMM \cite{pinyoanuntapong2023mmm}&
      CVPR 2024 &
$\underline{0.515}^{\pm.002}$ &
$\underline{0.708}^{\pm.002}$ &
$\underline{0.804}^{\pm.002}$ &
$\textbf{0.089}^{\pm.005}$ &
$\underline{2.926}^{\pm.007}$ &
$9.577^{\pm.050}$ &
$1.226^{\pm.035}$ \\
  &MotionDiffuse \cite{zhang2022motiondiffuse}&
      TPAMI 2024 &
    $0.491^{\pm.001}$ &
  $0.681^{\pm.001}$ &
  ${0.782}^{\pm.001}$ &
  $0.630^{\pm.001}$ &
  $3.113^{\pm.001}$ &
  $9.410^{\pm.049}$ &
  $1.553^{\pm.042}$ \\
            &GUESS \cite{gao2024guess}&
                TVCG 2024 &
    $0.503^{\pm.003}$ &
   $0.688^{\pm.002}$ &
   $0.787^{\pm.002}$ &
     $0.109^{\pm.007}$ &
      $3.006^{\pm.007}$ &
      ${9.826}^{\pm.104}$ &
       $2.430^{\pm.100}$ \\       
\cmidrule(l){2-10}
   &MOST&
       - &
$ \textbf{0.526}^{\pm.006}$ &
$ \textbf{0.715}^{\pm.004}$ &
$ \textbf{0.810}^{\pm.004}$&
  $\underline{0.092}^{\pm.012}$&
$ \textbf{2.865}^{\pm.019}$ &
$\underline{9.555}^{\pm.092}$ &
$ \textbf{2.821}^{\pm.120}$
   \\ 
   \midrule

&Real &
    - &
  $0.424^{\pm.005}$ &
  $0.649^{\pm.006}$ &
  $0.779^{\pm.006}$ &
  $0.031^{\pm.004}$ &
  $2.788^{\pm.012}$ &
  $11.08^{\pm.097}$ &
  \multicolumn{1}{c}{-}
  \\

\multirow{15}{*}{KIT-ML}&TEMOS \cite{petrovich2022temos}&
    ECCV 2022&
  $0.353^{\pm.006}$ &
  $0.561^{\pm.007}$ &
    $0.687^{\pm.005}$ & 
    $3.717^{\pm.051}$ & 
    $3.417^{\pm.019}$ & 
    $10.84^{\pm.100}$ & 
    $0.532^{\pm.034}$ \\
&Temporal VAE \cite{guo2022generating}&
    CVPR 2022 &
  $0.361^{\pm.006}$ &
  $0.559^{\pm.007}$ &
  $0.693^{\pm.007}$ &
  $2.770^{\pm.109}$ &
  $3.401^{\pm.008}$ &
  $10.91^{\pm.119}$ &
  $1.482^{\pm.065}$ \\
  &MLD \cite{chen2023executing}&
      CVPR 2023 &
    $0.390^{\pm.008}$ &
  $0.609^{\pm.008}$ &
  $0.734^{\pm.007}$ &
  $0.404^{\pm.027}$ &
  $3.204^{\pm.027}$ &
  $10.80^{\pm.117}$ &
  $2.192^{\pm.071}$ \\
        &T2M-GPT \cite{zhang2023generating}&
            CVPR 2023 &
    $0.416^{\pm.006}$ &
  $0.627^{\pm.006}$ &
   $0.745^{\pm.006}$ &
    $0.514^{\pm.029}$ &
     $3.007^{\pm.023}$ &
      $10.92^{\pm.108}$ &
       $1.570^{\pm.039}$ \\  
  &MDM \cite{tevet2022human}&
      ICLR 2023 &
    $0.164^{\pm.004}$ &
  $0.291^{\pm.004}$ &
  $0.396^{\pm.004}$ &
  $0.497^{\pm.021}$ &
  $9.190^{\pm.022}$ &
  $10.85^{\pm.109}$ &
  $1.907^{\pm.214}$ \\

          &Fg-T2M \cite{wang2023fg}&
              ICCV 2023 &
    $0.418^{\pm.005}$ &
  $0.626^{\pm.004}$ &
   $0.745^{\pm.004}$ &
    $0.571^{\pm.047}$ &
     $3.114^{\pm.015}$ &
      $10.93^{\pm.083}$ &
       $1.019^{\pm.029}$ \\

            &ReMoDiffuse \cite{zhang2023remodiffuse}&
                ICCV 2023 &
    $0.427^{\pm.014}$ &
   $0.641^{\pm.004}$ &
   $0.765^{\pm.055}$ &
     $\underline{0.155}^{\pm.006}$ &
      $2.814^{\pm.012}$ &
      $10.80^{\pm.105}$ &
       $1.239^{\pm.028}$ \\
       
  &MotionGPT \cite{jiang2024motiongpt}&
      NeurIPS 2024 &
  $0.366^{\pm.005}$ &
  $0.558^{\pm.004}$ &
  $0.680^{\pm.005}$ &
  $0.510^{\pm.016}$ &
  $3.527^{\pm.021}$ &
  $10.35^{\pm.084}$ &
   $\underline{2.328}^{\pm.117}$ \\
&FineMoGen \cite{zhang2024finemogen}&
    NeurIPS 2024 &
$\underline{0.432}^{\pm.006}$ &
$\underline{0.649}^{\pm.005}$ &
$\underline{0.772}^{\pm.006}$ &
$0.178^{\pm.007}$ &
$2.869^{\pm.014}$ &
$10.85^{\pm.115}$ &
$1.877^{\pm.093}$ \\
  &MMM \cite{pinyoanuntapong2023mmm}&
      CVPR 2024 &
$0.404^{\pm.005}$ &
$0.621^{\pm.005}$ &
$0.744^{\pm.004}$ &
$0.316^{\pm.028}$ &
$2.977^{\pm.019}$ &
$10.91^{\pm.101}$ &
$1.232^{\pm.039}$ \\
       &MotionDiffuse \cite{zhang2022motiondiffuse}&
    TPAMI 2024 &
  $0.417^{\pm.004}$ &
  $0.621^{\pm.004}$ &
  $0.739^{\pm.004}$ &
  $1.954^{\pm.062}$ &
  $2.958^{\pm.005}$ &
  $\textbf{11.10}^{\pm.143}$ &
  $0.730^{\pm.013}$ \\
            &GUESS \cite{gao2024guess}&
                TVCG 2024 &
   $0.425^{\pm.005}$ &
   $0.632^{\pm.007}$ &
   $0.751^{\pm.005}$ &
     $0.371^{\pm.020}$ &
      $ \textbf{2.421}^{\pm.022}$ &
      $10.93^{\pm.110}$ &
       $\textbf{2.732}^{\pm.084}$ \\
\cmidrule(l){2-10}

       &MOST&
           - &
    $\textbf{0.436}^{\pm.007}$ &
  $\textbf{0.658}^{\pm.004}$ &
   $\textbf{0.783}^{\pm.008}$ &
    $\textbf{0.139}^{\pm.019}$ &
     $\underline{2.732}^{\pm.024}$ &
      $\underline{11.01}^{\pm.093}$ &
       $1.739^{\pm.136}$ \\
\midrule
\end{tabular}%
}
\caption{Text-to-motion generation results on the HumanML3D~\cite{guo2022generating} and KIT-ML~\cite{plappert2016kit} datasets. For each metric, we repeat the evaluation 20 times and report the average with a 95\% confidence interval.
`$\uparrow$'(`$\downarrow$') indicates that the values are better if the metric is larger (smaller). `$\rightarrow$' means the closer to real motion the better. The best and the second-best results are bolded and underlined, respectively. }
\label{compare_humanml3d}
\end{table*}

\subsubsection{Implementation Details}
In the T2M retrieval phase, a 4-layer transformer is adopted for the motion decoder, while the text encoder incorporates CLIP along with an additional 2-layer transformer. The setting of $\lambda_\text{B}$ at 0.5 is motivated by its influence on the overall process. For training, we randomly choose one text from a motion to serve as the corresponding item. For testing, we use the initial text from each motion. To ensure diversity, we remove samples with identical texts. We employ the Adam optimizer with a batch size of 128, learning rate of 5e-5, and a temperature $\tau$ of 0.1. The embedding dimensionality is set to d = 512.

In T2M generation using the diffusion model, we employ 1000 diffusion steps and linearly adjust the variance $b_t$ within the range of 0.0001 to 0.02. Adam optimization is carried out with a batch size of 1024. Consistency is maintained between the motion transformer and text prompt encoder setups used in T2M retrieval. Hyperparameters $T_s$ (set to 5), $K_e$ (set to 10), and $n$ (set to 2) are carefully chosen for optimization. The embedding dimensionality remains $d = 512$, with the latent dimensionality of the feedforward layer set to 2048. In the classifier-free setting, dropout probability is 10\%, and the scale hyperparameter $s$ is adjusted to 3.2 in KIT-ML \cite{plappert2016kit} and 4.7 in HumanML3D \cite{guo2022generating}. To ensure fairness, each run is conducted 20 times, with results reported using a 95\% confidence interval.

To train the temporal clip Banzhaf interaction predictor, we utilize MSE loss, allowing the model to learn Banzhaf mappings for input matrices. It takes a text-motion similarity matrix and predicts the corresponding Banzhaf Interaction. This model architecture includes five linear layers, four ReLU activation functions, and a self-attention module.

Pose representation follows Guo \cite{guo2022generating}, consisting of seven parts: ($r^{va}, r^{vx}, r^{vz}, r^{h}, j^{p}, j^{v}, j^{r}$). Here, $r^{va} \in \mathbb{R}$ represents the root joint's angular velocity along the Y-axis, $r^{vx}, r^{vz} \in \mathbb{R}$ denote the root joint's linear velocities along the X and Z axes. $r^h \in \mathbb{R}$ indicates the root height. $j^p, j^v \in \mathbb{R}^{J\times3}$ denote the positions and linear velocities of each joint, while $j^r \in \mathbb{R}^{J\times6}$ represents the 6D rotation of each joint. Here, $J$ signifies the number of joints, 22 for the HumanML3D \cite{guo2022generating} dataset and 21 for the KIT-ML \cite{plappert2016kit} dataset.

\begin{table}[t]
\centering

\begin{tabular}{>{\raggedright\arraybackslash}p{0.2\linewidth} >{\raggedright\arraybackslash}p{0.7\linewidth}}
\toprule
\textbf{Level of Rareness} & \textbf{Sample Prompts} \\
\midrule
Tail 0-5\% &-  A person does a single knee down with left leg with right leg stepping forward while raising right arm up to head level and placing its forearm in front of face, and then resumes the original position. \\
 &- The man makes two spartan kicks, followed by two medium high kicks.  \\
\midrule
Tail  0-15\%&- This person zig zags forward then stops to the right. \\ 
 &- A man does a push up and then uses his arms to balance himself back to his feet. \\
\midrule
Tail 0-25\%&- A person lifts something to their face and wobbles their body in circles. \\
 &- A figure jazzily steps backward then walks forward, as though walking to music or a beat. \\
 \midrule
Tail 75-100\%&- The person is walking back and forth. \\
 &- A person picks something up. \\
\bottomrule

\end{tabular}
\caption{Sample prompts showcasing different rareness within the descriptions.}
\label{rareness_tab}
\end{table}

\subsection{Comparison with State-of-the-arts}

\subsubsection{Evaluation on Retrieval}
We compared MOST with various advanced T2M retrieval models in Table \ref{retrieval_h3d_kit}, including those mentioned in \cite{petrovich2022temos,guo2022generating,petrovich23tmr}. 
The evaluation protocol of ``All'' is the all test set that is used. The evaluation protocol of ``Small Batches'' is that we randomly select batches of 32 motion-text pairs and report their average performance.
Notably, we have achieved significant improvements in both the KIT-ML~\cite{plappert2016kit} and HumanML3D~\cite{guo2022generating} datasets. Our results demonstrate substantial progress in enhancing R@k and MedR metrics, indicating that our approach excels in comprehending the intricate connections between the provided text prompts and the corresponding motion sequences. 
This model capability enables us to discover motion clips that closely align with the intended meaning of the text.

\begin{table*}[t]
\centering
\resizebox{\linewidth}{!}{
\begin{tabular}{l|c|ccc|ccc|ccc|cc}
\toprule
\multirow{3}{*}{Methods} & \multirow{3}{*}{Datasets}& \multicolumn{3}{c|}{Rare Texts } & \multicolumn{3}{c|}{Rare Texts} & \multicolumn{3}{c|}{Rare Texts} & \multicolumn{2}{c}{Common Texts}\\

~ & &  & Tail 0-5\% Data&  &  & Tail 0-15\% Data &  &  & Tail 0-25\% Data &  & \multicolumn{2}{c}{Tail 75-100\% Data} \\
~ & & FID$\downarrow$ & MM$\downarrow$ & W-MM$\downarrow$ & FID$\downarrow$ & MM$\downarrow$ & W-MM$\downarrow$ & FID$\downarrow$ & MM$\downarrow$ & W-MM$\downarrow$ & FID$\downarrow$ & MM$\downarrow$ \\
\midrule

Temporal VAE\cite{guo2022generating}&
\multirow{5}{*}{HumanML3D}&
  $3.55$ &
  $3.89$ &
  $21.02$ &
  $2.36$ &
  $3.91$ &
  $16.56$ &
  $2.18$ &
$3.85$ &
  $15.23$ &
$0.59$ &
  $2.85$ 
  \\
  MotionDiffuse\cite{zhang2022motiondiffuse}&
   ~&
  $1.48$&
  $3.77$ &
$19.81$ &
$1.32$ &
$3.54$ &
$15.29$ &
$1.19$ &
$3.46$ &
$13.68$&
$0.68$ &
  $2.77$  
  \\
    Fg-T2M \cite{wang2023fg}&
    ~&
   $1.26$&
   $3.70$&
    $19.49$&
    $1.17$&
    $3.40$&
  $15.12$&
     $0.98$&
     $3.37$&
    $13.03$&
$0.54$ &
  $2.64$ 
  \\
      FineMoGen \cite{zhang2024finemogen}&
    ~&
   $0.99$&
   $3.63$&
    $19.02$&
    $0.49$&
    $3.36$&
  $13.03$&
     $0.32$&
     $3.33$&
    $12.91$&
$0.29$ &
  $2.56$ 
  \\
  ReMoDiffuse\cite{zhang2023remodiffuse}&
  ~&
  $0.87$&
  $3.51$ &
$18.71$ &
$0.43$ &
$3.29$ &
$12.76$ &
$0.39$ &
$3.31$ &
$12.83$ &
$0.18$ &
  $2.50$ 
  \\
MOST &

~&
$\textbf{0.66}$&
$\textbf{3.39}$ &
$\textbf{16.80}$ &
$\textbf{0.34}$ & 
$\textbf{3.21}$ &
$\textbf{12.26}$ &
$\textbf{0.19}$ &
$\textbf{3.25}$ &
$\textbf{11.78}$ &
$\textbf{0.14}$ &
$\textbf{2.39}$ 
\\

\midrule

Temporal VAE\cite{guo2022generating}&
\multirow{5}{*}{KIT-ML}&
  $23.81$ &
  $6.56$ &
  $38.02$ &
  $8.54$ &
  $4.63$ &
  $24.26$ &
  $6.97$ &
$4.69$ &
  $19.03$&
$1.15$ &
  $2.30$ 
  \\

  MotionDiffuse\cite{zhang2022motiondiffuse}&
  ~&
  $21.54$ &
  $5.93$ &
  $35.50$ &
  $8.22$ &
  $4.39$ &
  $23.97$ &
  $6.58$&
    $4.18$ &
  $18.50$ &
$0.72$ &
  $2.10$ 
  \\
  
        Fg-T2M \cite{wang2023fg}&
          ~&
   $15.67$&
   $5.71$&
    $33.12$&
    $5.90$&
    $4.30$&
  $23.73$&
     $3.55$&
     $4.11$&
     $18.09$&
$0.66$ &
  $2.03$ 
     \\
           FineMoGen \cite{zhang2024finemogen}&
    ~&
   $8.92$&
   $5.64$&
    $32.51$&
    $1.57$&
    $4.16$&
  $22.57$&
     $1.60$&
     $3.93$&
    $17.43$&
$0.39$ &
  $2.03$ 
  \\
    ReMoDiffuse\cite{zhang2023remodiffuse}&
      ~&
  $8.57$ &
  $5.44$ &
  $31.75$ &
  $1.15$ &
  $3.97$ &  
  $22.09$ &
  $0.43$&
    $3.94$ &
  $17.58$ &
$0.40$ &
  $2.04$ 
  \\

MOST &
  ~&
  $\textbf{7.34}$&
  $\textbf{4.82}$ &
$\textbf{29.92}$ &
$\textbf{0.96}$ &
$\textbf{3.91}$ &
$\textbf{21.21}$ &
$\textbf{0.31}$ &
$\textbf{3.71}$ &
$\textbf{16.64}$ &
$\textbf{0.26}$ &
$\textbf{1.97}$ 
\\

\bottomrule
\end{tabular}%
 }

\caption{Evaluation of generalization ability on HumanML3D~\cite{guo2022generating} and KIT-ML~\cite{plappert2016kit} datasets. We report the best results in bolded.}
\label{generalization}
\end{table*}

\subsubsection{Evaluation on Generation}

Table \ref{compare_humanml3d} compares our approach with SOTA T2M generation models, including \cite{petrovich2022temos,guo2022generating,chen2023executing,zhang2023generating,tevet2022human,wang2023fg,zhang2023remodiffuse,jiang2024motiongpt,zhang2024finemogen,zhang2022motiondiffuse,gao2024guess}. Notably, our model exhibits exceptional FID performance on the KIT-ML \cite{plappert2016kit} dataset compared to MMM \cite{pinyoanuntapong2023mmm}, which struggled on this dataset due to its small scale. Similarly, while GUESS \cite{gao2024guess} achieved reasonable MM-Dist on KIT-ML, it faced challenges with the more complex HumanML3D \cite{guo2022generating} datasets. Compared to ReMoDiffuse \cite{zhang2023remodiffuse}, our approach demonstrates superior precision metrics (R-TOP, FID, and MM Dist) thanks to the utilization of retrieved motion clips, effectively mitigating motion redundancy and enhancing both matching accuracy and diversity. However, it is important to note that diversity metrics (MultiModality and Diversity) become less relevant if the generated motions do not align with the expected results.

\begin{figure}[t]
    \centering
     \includegraphics[width=\linewidth]{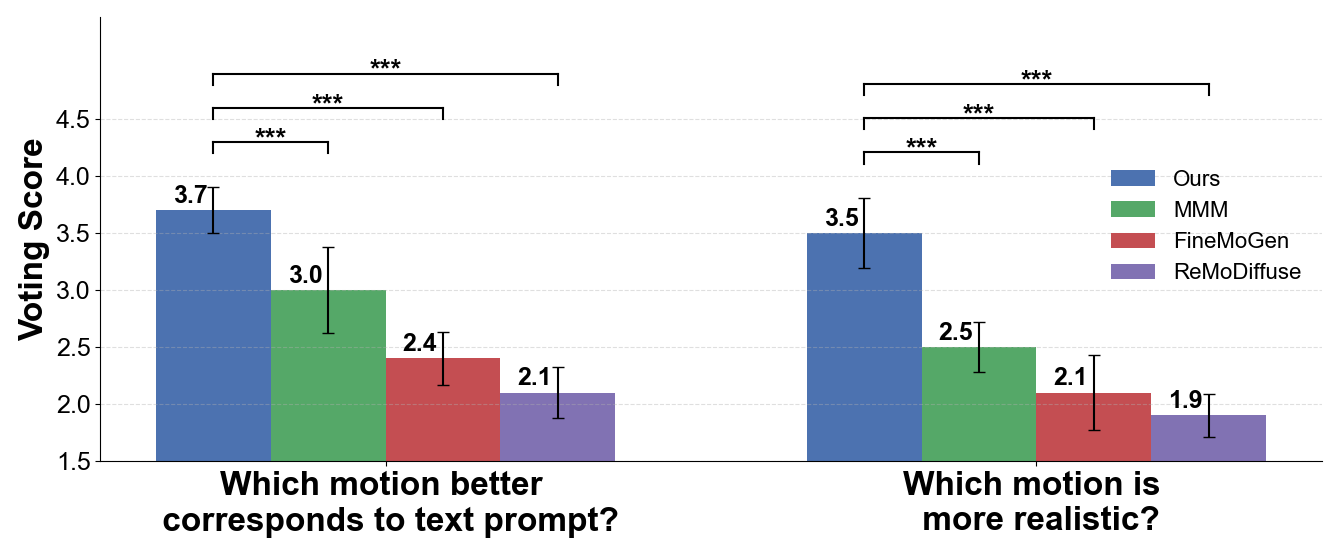}
    \caption{The result of user study. Each bar represents the voting scores of methods, with higher values being better. The voting score ranges from 1 to 4.
    }
    \label{usestudy}
\end{figure}

\textit{User Study}: We also evaluated performance through a user study, comparing MOST with FineMoGen \cite{zhang2024finemogen}, ReMoDiffuse \cite{zhang2023remodiffuse}, and MMM \cite{pinyoanuntapong2023mmm}. This study involved 30 participants (aged 23-28 years, 27 male and 3 female), including 10 animation researchers, 2 of whom were familiar with our method, and 20 general users. Prior to the experiment, all participants received a 10-minute training session to understand the motion-text alignment task and evaluation criteria.
Participants voted on two questions: ``Which of the two motions corresponds better to the text prompt?'' and ``Which of the two motions is more realistic?'' The statistical results of the user study, presented in Figure \ref{usestudy}, indicate that our MOST method outperformed state-of-the-art approaches on both questions, demonstrating its superior ability to generate motions from rare texts. Notably, our method achieved lower standard deviations, reflecting better stability.
Furthermore, we conducted paired t-tests. For each question, we drew a line from MOST to each of the other three methods and placed * to denote p<0.1, ** for p<0.01, and *** for p<0.001. The results revealed significant differences between MOST and each of the other three methods for both questions.


Overall, our method consistently performs favorably compared to other models across accuracy and diversity metrics, indicating its effectiveness in generating high-quality motion sequences that faithfully reflect the intended meaning of textual prompts. These results showcase the reliability of our approach in producing realistic and diverse motion outputs.

\begin{figure}[t]
    \centering
     \includegraphics[width=\linewidth]{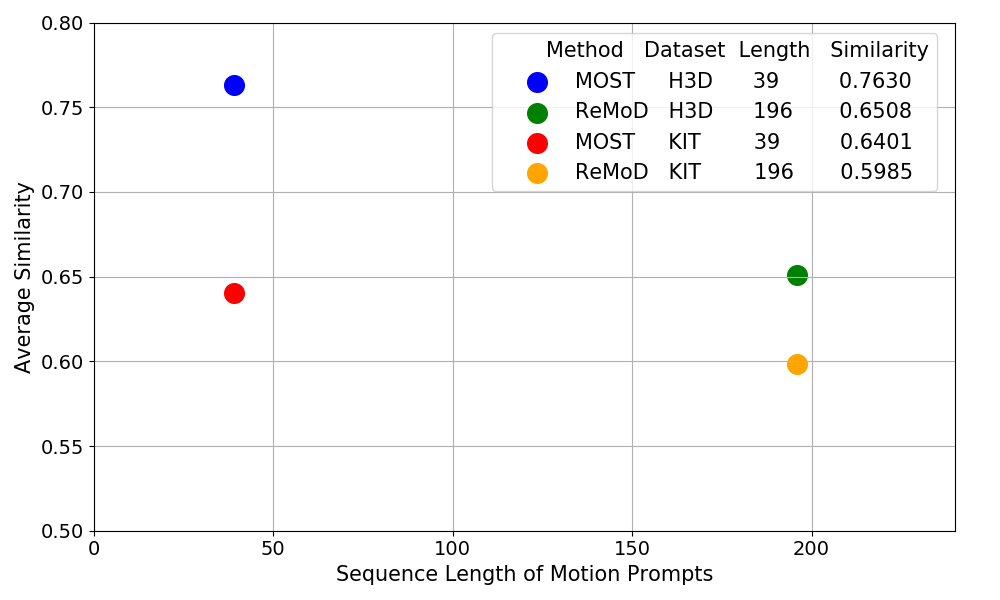}
    \caption{The experiments assessing motion redundancy issue. Visualization comparison of the sequence length of motion prompts and average motion feature similarity. The closer the model is to the upper left corner the better handling motion redundancy. ``H3D'' represents the HumanML3D dataset.
    }
    \label{motionreden}
\end{figure}

\subsubsection{Evaluation on Generalization}
In order to assess our model's adaptability to rare text, we have devised an experiment for evaluating generalization. To achieve this, we extended the metrics inspired by ReMoDiffuse \cite{zhang2023remodiffuse}, which introduces the concept of "Rareness" in prompts. To calculate the rareness $r_p$ of a test prompt $p$, we use the formula: 
\begin{equation}
 r_p = 1 - \max_i \{<E(\mathrm{text}_{i}), E(\mathrm{prompt})>\} 
\end{equation}
Here, $E$ signifies the CLIP text encoder, $\mathrm{text}_i$ is the text prompt, and $<\cdot,\cdot>$ represents cosine similarity. This formula quantifies the maximum similarity between the given prompt and the prompts in the training set. Higher similarity corresponds to lower levels of rareness, indicating a common text, and vice versa. We then sort all prompts in increasing level of rareness and select Tail 0-5\%, 0-15\%, and 0-25\% rare prompts, which correspond to 207, 623, and 1046 prompts, respectively.
We present two examples for each level of rareness to demonstrate the spectrum of complexity, as depicted in Table \ref{rareness_tab}. These samples are used to evaluate generalization through metrics like FID, MM Dist, and Weight-MM (W-MM).
We propose the W-MM for text prompt $j$ as:
\begin{equation}
\mathrm{W}\text{-}\mathrm{MM}^j =\frac{\beta r_p^j \mathrm{MM}^j}{\max (r_p)}
\end{equation}
Here, $r_p^j$ denotes the rareness of text prompt $j$, $\max (r_p)$ is the maximum rareness in the dataset, $\mathrm{MM}^j$ represents MM Dist for $j$, and $\beta$ is a control coefficient set to 10. W-MM assigns greater weight to rarer texts, amplifying their impact in calculating W-MM.

\begin{table}[t]
\centering
\large
\resizebox{\columnwidth}{!}{%
\begin{tabular}{llllllllllll}
\toprule
\multicolumn{4}{l}{\cellcolor{lightgray}\textbf{Ablation Analysis}} & & \multicolumn{2}{l}{All Texts (Tail 0-100\% Data)}&  & \multicolumn{2}{l}{Rare Texts (Tail 0-15\% Data)}  &         \\ \cmidrule(lr){1-4}\cmidrule(lr){6-7} \cmidrule(l){9-10}  

                        R
                        & S 
                        & N
                        & M
                        &
                        & FID $\downarrow$ 
                        & MM $\downarrow$
                        &  
                        & FID $\downarrow$
                        &MM $\downarrow$

                        \\ \midrule
$-$ & $-$ & $-$ & $-$  &                                & $0.61$ & $3.20$   &  & $8.78$& $4.21$   \\
Base & $-$ & $2$ &Entire Motion   &      & $0.40$ & $3.10$   &  & $3.81$& $3.95$   \\
Banzhaf & $4$ & $2$ & Motion Clip &     & $0.27$ & $2.89$   &  & $1.30$& $4.04$   \\
Banzhaf& $5$ & $2$ & Motion Clip  &      & $0.13$ & $2.79$   &  & $0.96$& $3.91$    \\
Banzhaf& $5$ & $3$ & Motion Clip  &      & $0.23$ & $2.98$   &  & $1.35$& $3.95$   \\
Banzhaf& $5$ & $4$ & Motion Clip  &      & $0.43$ & $3.12$   &  & $1.77$& $4.11$    \\
Banzhaf & $6$ & $2$ &Motion Clip  &     & $0.31$ & $2.99$   &  & $1.69$& $3.97$   \\
\midrule

\multicolumn{4}{l}{\cellcolor{lightgray}\textbf{Component Analysis}}&      &  &   &  & & \\
\midrule
\multicolumn{4}{l}{MotionDiffuse \cite{zhang2022motiondiffuse}}&      & $1.95$ & $2.95$   &  & $8.22$& $4.39$\\
\multicolumn{4}{l}{MotionDiffuse + Entire Motion}&      & $0.87 (\textcolor{darkgreen}{\downarrow55\%})$ & $2.90 (\textcolor{darkgreen}{\downarrow1.7\%})$   &  & $2.08 (\textcolor{darkgreen}{\downarrow74\%})$& $4.17 (\textcolor{darkgreen}{\downarrow5.0\%})$\\

\multicolumn{4}{l}{MotionDiffuse + Motion Clip}      & & $0.56 (\textcolor{red}{\downarrow71\%})$ & $2.87 (\textcolor{red}{\downarrow2.7\%})$   &  & $1.34 (\textcolor{red}{\downarrow83\%})$& $4.05 (\textcolor{red}{\downarrow7.7\%})$\\
\midrule
\multicolumn{4}{l}{Fg-T2M \cite{wang2023fg}}  &      & $0.57$ & $3.11$   &  & $5.90$& $4.08$\\
\multicolumn{4}{l}{Fg-T2M + Entire Motion}&      & $0.29 (\textcolor{darkgreen}{\downarrow49\%})$ & $2.93 (\textcolor{darkgreen}{\downarrow5.7\%})$   &  & $2.23 (\textcolor{darkgreen}{\downarrow62\%})$& $4.01 (\textcolor{darkgreen}{\downarrow1.7\%})$\\

\multicolumn{4}{l}{Fg-T2M + Motion Clip}      & & $0.21 (\textcolor{red}{\downarrow63\%})$ & $2.84 (\textcolor{red}{\downarrow8.6\%})$   &  & $1.21 (\textcolor{red}{\downarrow79\%})$& $3.94 (\textcolor{red}{\downarrow3.4\%})$\\
\midrule
\multicolumn{4}{l}{MOST (concat) }  &      & $0.19$ & $2.83$   &  & $1.23$& $3.99$\\
\multicolumn{4}{l}{MOST (cross-attention)}   & & $0.13$ & $2.79$   &  & $0.96$& $3.91$\\

\bottomrule
\end{tabular}%
 }
\caption{Ablation analysis and component analysis on KIT-ML~\cite{plappert2016kit} dataset. ``R'' denotes the retrieval strategy. ``Banzhaf'' and ``Base'' represents with and without Banzhaf interaction, respectively. ``S'' is the motion clips number. ``N'' indicates the prompt motions number. ``M'' indicates the utilization of motion clips or entire motion in motion prompt module.}
\label{ablation}
\end{table}

Table \ref{generalization} compares our approach with state-of-the-art T2M generation models. In comparison to \cite{guo2022generating}, \cite{wang2023fg}, \cite{zhang2022motiondiffuse} and \cite{zhang2024finemogen}, we outperform current methods across all metrics for prompts with rareness in Tail 0-5\%, 0-15\%, and 0-25\%. This highlights the generalizability of our proposed prompt-generation framework, showcasing its robustness in generating motion even under conditions of rare text. In contrast to the text-text retrieval-based method \cite{zhang2023remodiffuse}, we demonstrate significant improvement for the Tail 0-5\% prompts, indicating that integrating motion clips and motion prompt module information provides more valuable motion prompts to enhance the generation process. Overall, our MOST maintains high-quality generation even with rare and challenging texts, proving its remarkable generalization and robustness. Meanwhile, we have also conducted experiments on the HumanML3D \cite{guo2022generating} and KIT-ML's \cite{plappert2016kit} common text prompts (Tail 75-100\% prompts), as shown in Table \ref{generalization}, which demonstrates that MOST still has competitive performance under common text prompts.

\subsubsection{Evaluation on Assessing Motion Redundancy}
To assess the extent to which the issue of motion redundancy has been addressed, we conducted related experiments with ReMoDiffuse \cite{zhang2023remodiffuse} to verify our effectiveness, as shown in Fig. \ref{motionreden}. For a text $i$, we calculate the average Cosine similarity of the motion clips retrieved by MOST and ground truth motion data in the test sets, which are also calculated with the overall motion retrieved by ReMoDiffuse \cite{zhang2023remodiffuse}. We extracted the motion features using Guo's pre-trained motion feature extractor \cite{guo2022generating} and calculated the average cosine similarity by:
\begin{equation}
S_{ave} = \frac{ \sum_{i=1}^{z} Cos<F_{gt\_motion}^{i}, F_{motion\_prompt}^{i}>}{z}
\end{equation}
where $S_{ave}$ denotes average cosine similarity, $z$ represents the number of samples in the dataset, $F_{gt\_motion}^{i}$ and $ F_{motion\_prompt}^{i}$ are ground truth motion data and retrieved motion prompts corresponding to text $i$, respectively. The results show that, compared to ReMoDiffuse's  \cite{zhang2023remodiffuse} motion prompts of 196 frames, MOST only used the motion clips of 39 frames, which is only 20\% of the length, significantly alleviating the problem of motion redundancy. At the same time, it achieved a higher average cosine similarity than ReMoDiffuse \cite{zhang2023remodiffuse} on both the HumanML3D \cite{guo2022generating} and KIT-ML \cite{plappert2016kit} datasets, demonstrating that MOST accurately captures the core content of the motions through carefully selected motion clips. This indicates that MOST not only solves the problem of motion redundancy but also obtains higher-quality motion prompts to optimize and improve the motion generation process.

\begin{figure}[t]
    \centering
     \includegraphics[width=\linewidth]{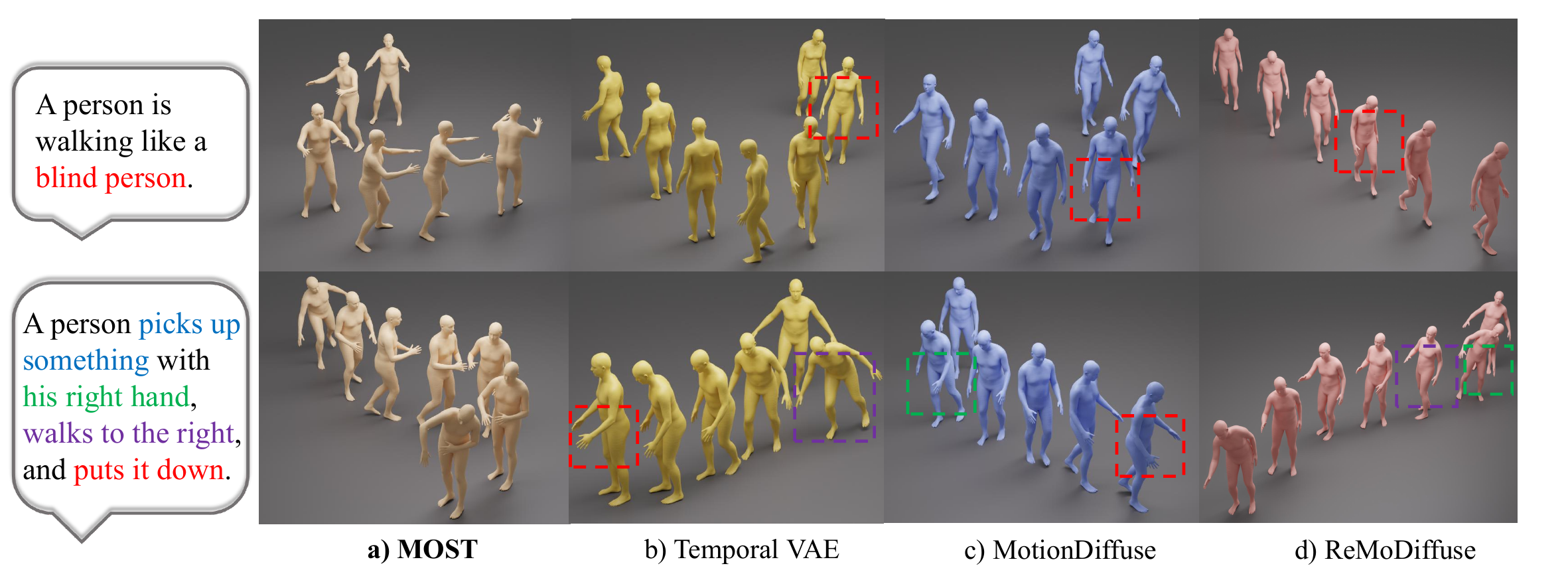}
    \caption{Visual results compared with existing methods. 
    The first line is the generated result on the rare text ``blind''. The second line is the generated result of complex action combinations.
    }
    \label{vis_compare}
\end{figure}

\begin{figure*}[t]
    \centering
    \begin{minipage}[t]{0.48\linewidth} 
        \centering
        \includegraphics[width=1.09\linewidth]{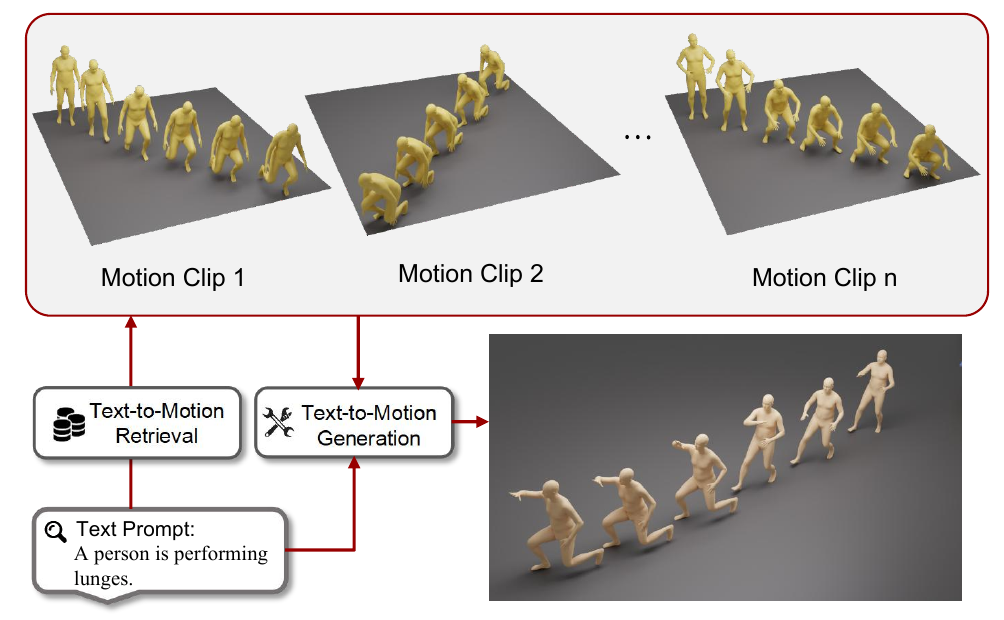}
        \vspace{-2.2em}
        \caption{Visualization on generation of the text prompt ``A person is performing lunges''.}
        \label{fig_vis1}
    \end{minipage}
    \hfill 
    \raisebox{1em}{
    \begin{minipage}[t]{0.48\linewidth} 
        \centering
        \includegraphics[width=\linewidth]{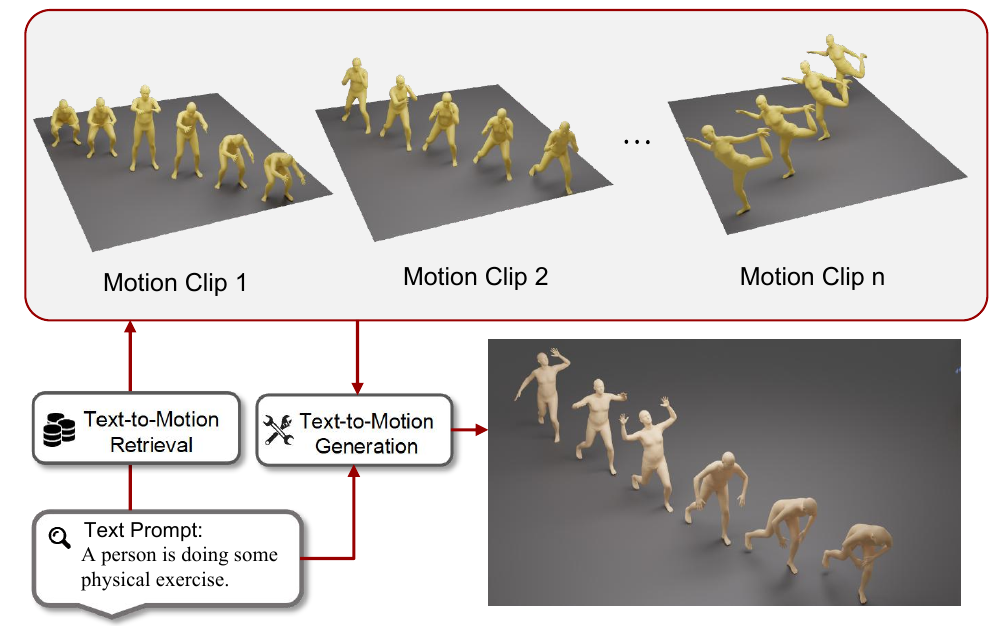}
        \caption{Visualization on generation of the text prompt ``A person is doing some physical exercise''.}
        \label{fig_vis2}
    \end{minipage}}
\end{figure*}
\begin{figure*}[t]
    \centering
    \begin{minipage}[t]{0.48\linewidth} 
    \centering
     \includegraphics[width=\linewidth]{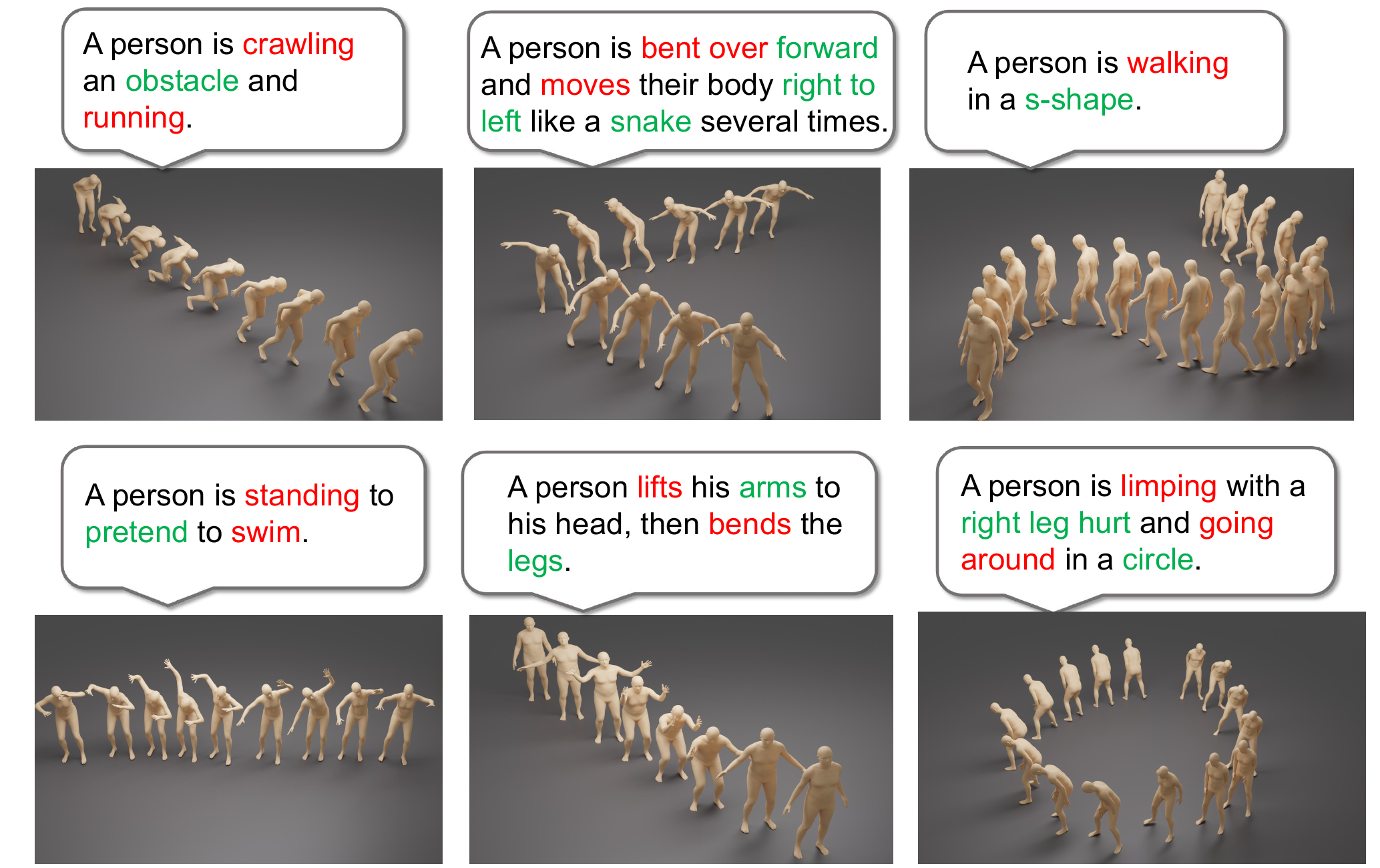}
             \vspace{-1.3em}
        \caption{More examples of visualizations. We mark the actions in red, and the detail descriptions in green.}
        \label{morevis}
    \end{minipage}
    \hfill 
    \raisebox{0.1em}{
    \begin{minipage}[t]{0.48\linewidth} 
        \centering
     \includegraphics[width=\linewidth]{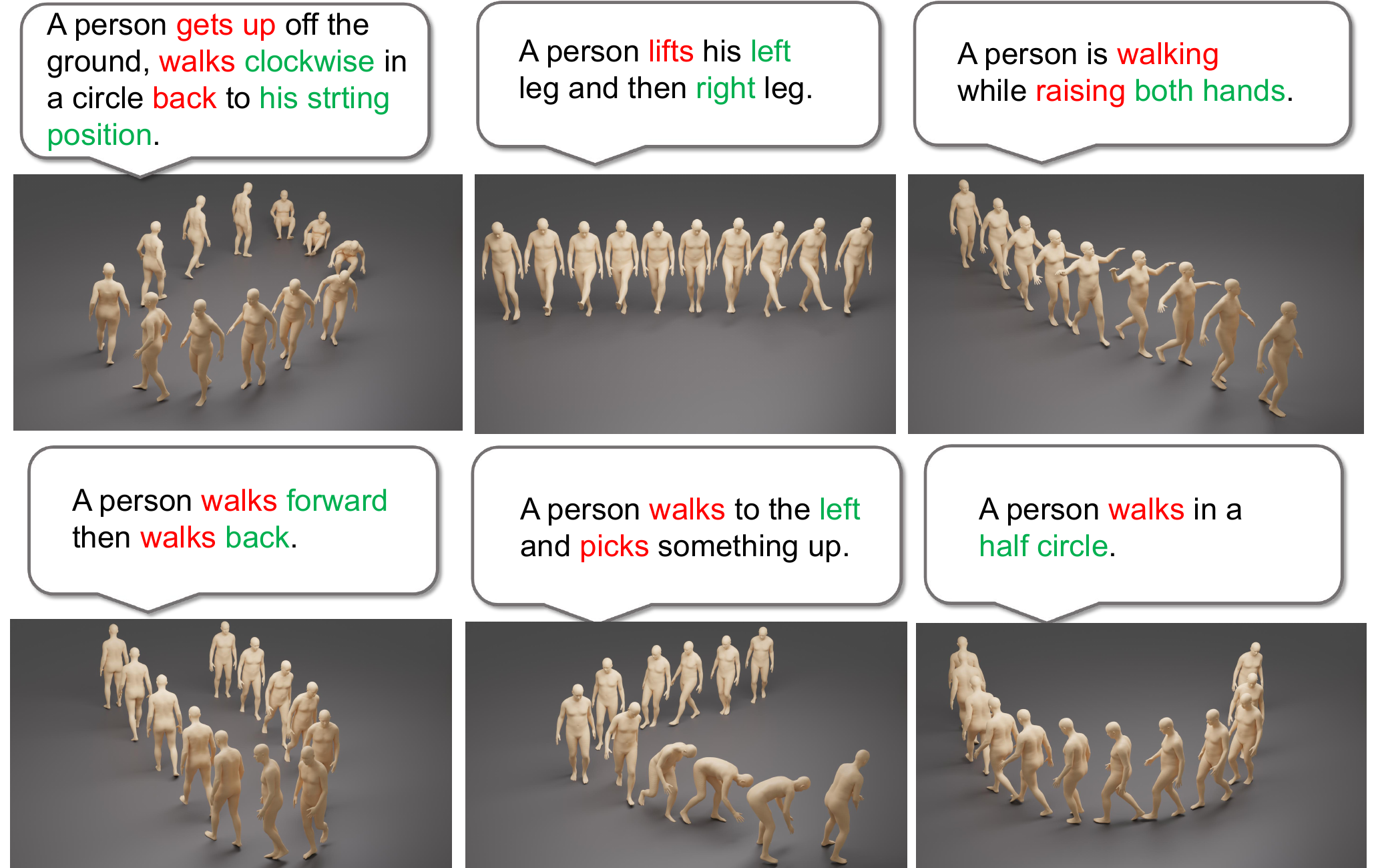}
        \caption{Visualizations in common texts. We mark the actions in red, and the detail descriptions in green.}
        \label{common}
    \end{minipage}}
\end{figure*}

\subsection{Component Analysis}

\subsubsection{The effectiveness of temporal clip Banzhaf interaction}

Existing datasets \cite{guo2022generating,plappert2016kit} primarily establish a general alignment between text and motion without detailed correspondence at the clip-level. This limitation hampers the ability of current methods to achieve fine-grained matching between text and motion clips. However, our temporal clip Banzhaf interaction method inherently assesses the correspondence between text and motion clips, effectively obtaining ground truth even without clip-level annotations. Leveraging this advantage, we proceed to analyze the effectiveness of temporal clip Banzhaf interaction in the following.

MOST and MOST (w/o Banzhaf) in Table \ref{retrieval_h3d_kit} highlight the enhancement brought by temporal clip Banzhaf interaction. In the T2M retrieval task, R@1 improves by 1\%, while median rank drops from 29 to 25 in the HumanML3D~\cite{guo2022generating} dataset. Furthermore,  Table \ref{ablation} illustrates the advantages of utilizing Banzhaf for retrieval strategy, significantly enhancing generation quality compared to the base method. This underscores the value of precise alignment through Banzhaf interaction between motion and text. We proceed to examine parameter impact in Table \ref{ablation}. 
Generally, a 39-frame (5 clips) motion captures key information. Excessive or insufficient clips lead to redundant or fragmented features. Excessive prompt motions can muddle denoising, potentially lowering accuracy. In contrast, utilizing motion clips in MOST yields superior performance in generation and generalization, effectively mitigating motion redundancy and providing evidence for the rationality of clip utilization.

\subsubsection{The effectiveness of retrieved motion clips}
To validate the benefits of our retrieved motion clips for motion generation and generalization tasks, we incorporate our retrieved motion clips and the motion prompt module into MotionDiffuse \cite{zhang2022motiondiffuse} and Fg-T2M~\cite{wang2023fg} to acquire external knowledge for retrieval. Additionally, we also compare the method of directly using entire motion sequences. The quantitative results in Table \ref{ablation} indicate that after learning retrieval knowledge, MotionDiffuse \cite{zhang2022motiondiffuse} and Fg-T2M~\cite{wang2023fg} generate motions that better align with real motion distributions and showcase robust generalization performance on the Tail 0-15\% of motion data. At the same time, compared with the method of using entire motion sequences, the performance using motion clips is superior, which also indicates that the motion redundancy brought by entire motion will pose challenges to the model's learning. This supports the effectiveness of our retrieval approach in enhancing motion generation.

\subsubsection{The impact of fusion mode}
We discuss the impact of fusion mode between the noise vector $X_t$ and the retrieved motion clips. Different fusion modes will affect the process of learning motion prompts. The common way in existing methods to fuse conditions into the denoiser is to concatenate or use the cross-attention mechanism. Therefore, we compare two ways using ``cross-attention'' and ``concat'' respectively in Table \ref{ablation}. 
For the ``cross-attention'' approach, we obtain motion prompt $R_i$ through the Motion Prompt Module. We design a cross-attention layer in the Motion Transformer, which learns the detailed relationship between the motion noise vector $X_t$ and the motion prompt $R_i$ to assist in motion generation. 
For the ``concat'' method, in Motion Transformer, we concat the motion noise vector $X_t$ with the motion prompt $R_i$ instead of designing a cross-attention layer. Employing MLPs to directly learn potential internal connections. 
From the experimental results, using ``cross-attention''  achieved better performance than the ``concat'' method in both FID and MM metrics, which indicated operating ``cross-attention'' can better grasp the intrinsic relationship between motion prompts to learn the fine-grained information more effectively.

\subsection{Qualitative Analysis}

The rare text prompts found in datasets like HumanML3D \cite{guo2022generating} and KIT-ML \cite{plappert2016kit} often involve unseen movements and complex action combinations. To assess the effectiveness of MOST in these scenarios, we compare it with SOTA methods in Fig. \ref{vis_compare}. The visualization highlights MOST's superior accuracy in portraying text prompts, particularly excelling in handling rare text instances. Notably, when confronted with unseen movements such as "blind person," ReModiffuse \cite{zhang2023remodiffuse}, FineMoGen \cite{zhang2024finemogen}, and Temporal VAE \cite{guo2022generating} struggle to interpret the rare term adequately, often capturing only generic actions like "walk." Furthermore, they face challenges in dealing with complex action combinations, often exhibiting errors in finer details like "left" and "right." In contrast, MOST adeptly performs these motions with greater fidelity, demonstrating its effectiveness in tackling such challenges. This robust performance in handling rare text scenarios underscores its strong generalization capabilities.

\begin{figure}[t]
    \centering
     \includegraphics[width=\linewidth]{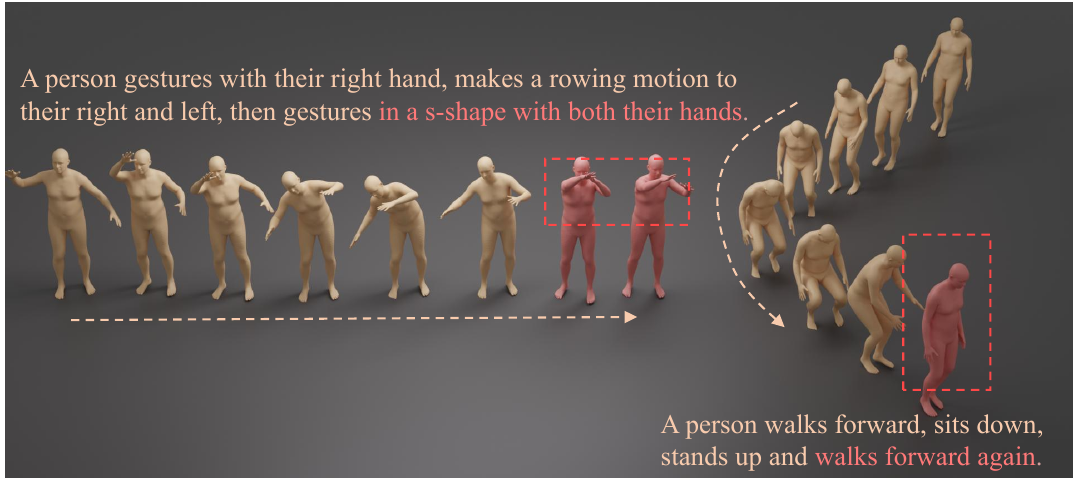}
    \caption{Visualization of some failure cases. The arrow represents the time axes and the red box indicates the incorrect motion frames.
    }
    \label{failure}
\end{figure}

\subsection{More Visualization on Generation} 

We begin with qualitative results showcasing the retrieval effectiveness of our method. As seen in Fig. \ref{fig_vis1} and \ref{fig_vis2}, the motion clips retrieved by our MOST align well with the text prompts. Retrieving motion clips most relevant to ``lunges" and ``physical exercise", benefits motion generation tasks and highlights our method's robust generalization capability. We also present additional generation effects of MOST across rare and common text scenarios, depicted in Fig. \ref{morevis} and \ref{common}. These examples demonstrate MOST-generated motions' superior alignment with text prompts, illustrating the remarkable performance of our architecture.
Finally, we provide some failure cases in Fig. \ref{failure}. While MOST effectively handles rare texts, it encounters challenges with longer, detailed texts, possibly due to difficulties in finely segmenting key clips within sentences, leading to suboptimal results.

Furthermore, we have presented more qualitative results under rare text conditions in the supplementary video, as well as side-by-side comparisons with state-of-the-art methods. The results clearly show that our MOST outperforms existing methods, validating the effectiveness of Temporal Clip Banzhaf Interaction in enhancing motion generation.

\begin{figure*}[h]
    \centering
    \begin{minipage}{0.49\textwidth}
        \centering
        \includegraphics[width=\linewidth]{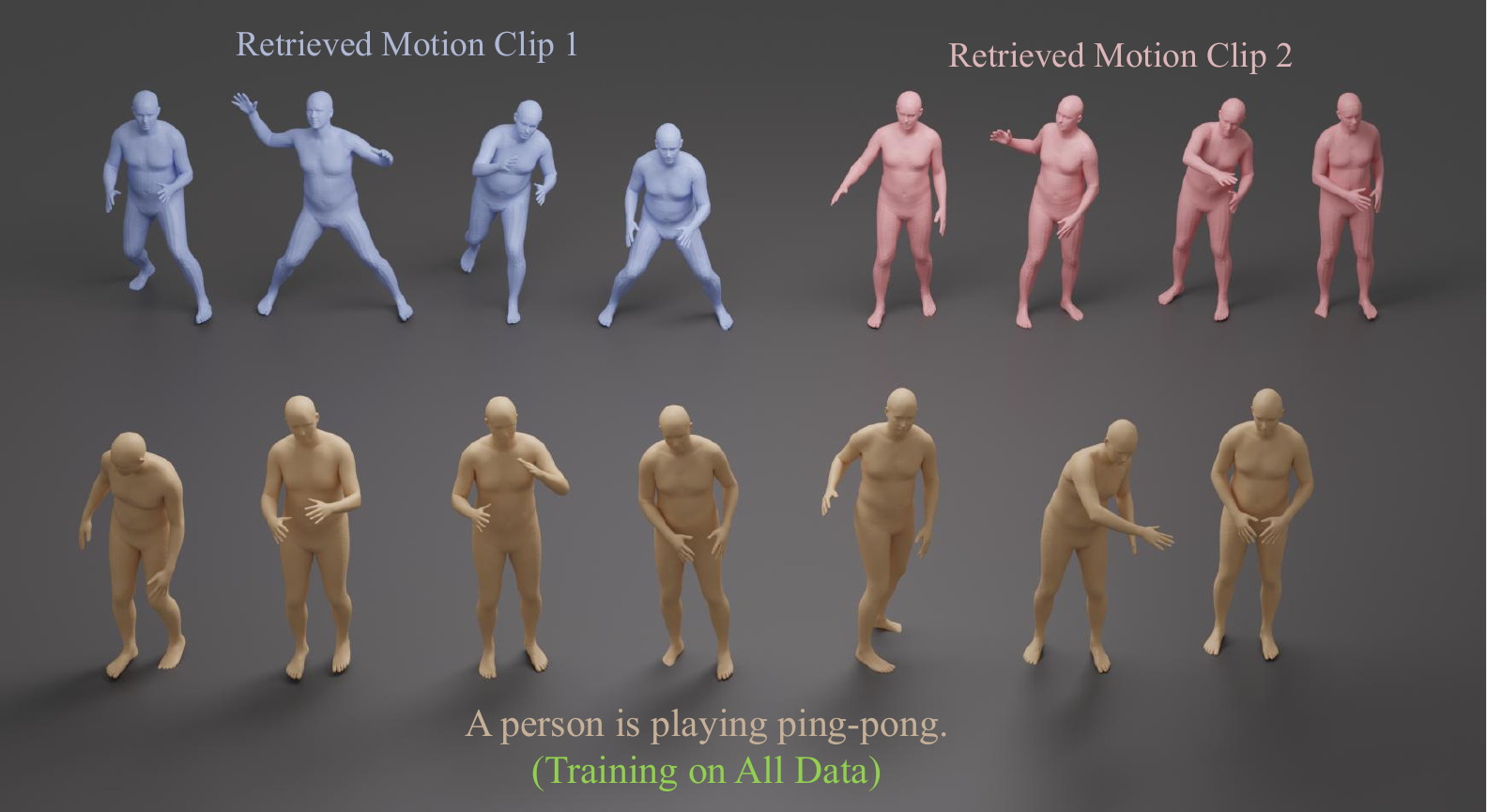}
    \end{minipage}
    \hfill
    \begin{minipage}{0.49\textwidth}
        \centering
        \includegraphics[width=\linewidth]{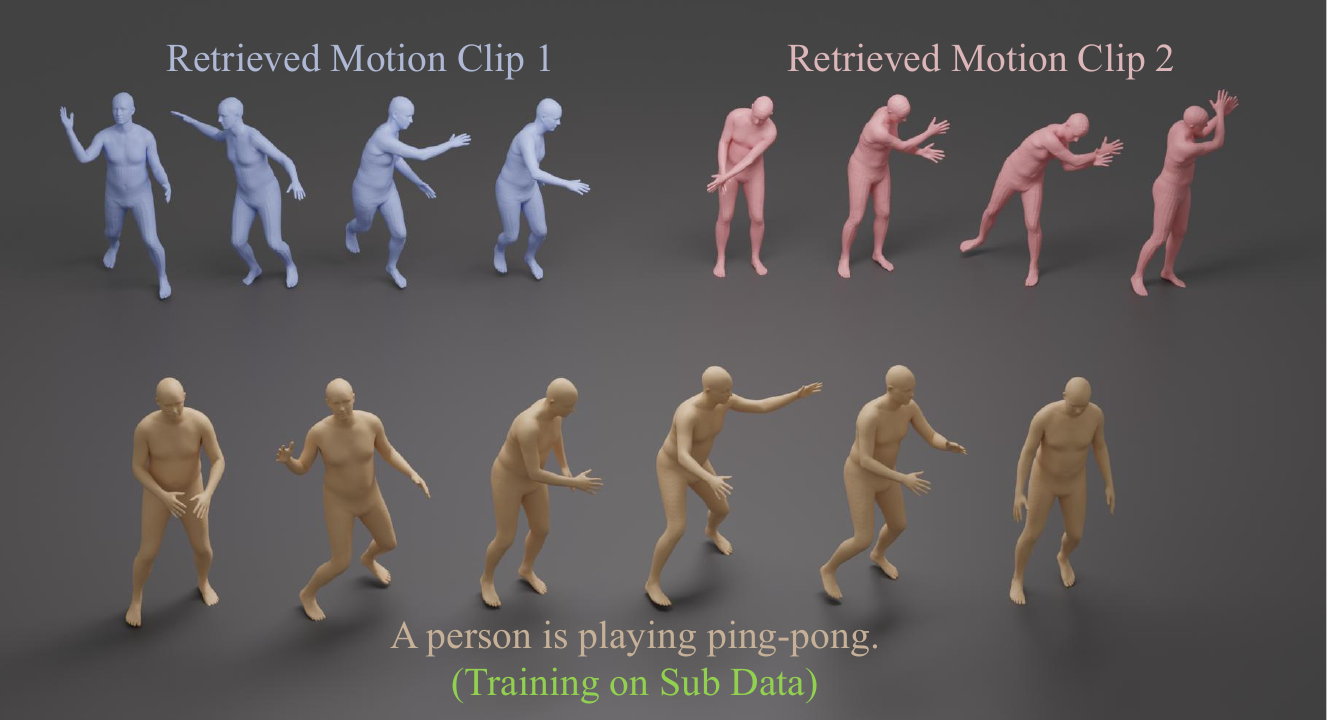}
    \end{minipage}
    \caption{The generated result for the sentence ``A person is playing ping-pong." The left figure shows the result trained on the all data, while the right figure shows the result trained on a sub-dataset with half of the data randomly dropped.}
    \label{fig:minor-pingpong}
\end{figure*}

\section{Limitations and Future Work}

We conducted an additional generalization experiment to study MOST’s retrieval and generation performance under varying dataset sizes and diversity. Specifically, we created a reduced dataset (Sub-HumanML3D) by randomly discarding half of the samples from HumanML3D, trained models on both datasets, and evaluated them on retrieval and generation tasks, as shown in Figure~\ref{fig:minor-pingpong}: ``A person is playing ping-pong.''

For the motion retrieval task, the model trained on the full dataset consistently retrieved motion clips containing the correct core elements (e.g., forward swinging). In contrast, the model trained on Sub-HumanML3D produced retrieval results that included forward swinging as well as golf-like motions, which are less semantically relevant to ping-pong. This comparison indicates that training on a larger dataset enables the model to retrieve more accurate and semantically aligned motion clips. Regarding the motion generation task, both models produced motions featuring forward swinging. However, only the model trained on the full HumanML3D dataset generated a ball-tossing motion. This suggests that a larger and more diverse dataset not only enhances retrieval accuracy but also improves the model’s generalization ability during generation. 
In summary, exploring how to better enhance the model’s generalization ability is a valuable direction for future work.

Additionally, we further discuss some limitations of MOST and suggest directions for future development. Firstly, its two-stage training and inference process can be inconvenient, hindering quick motion generation. A more streamlined, end-to-end model would enhance efficiency. Secondly, with the rapid advancement of large language models (LLMs), integrating them into MOST could improve motion generation, particularly for rare text conditions. By leveraging LLMs to interpret rare texts and generate more common ones, we can simplify subsequent retrieval and generation tasks. As an initial exploration of motion clip retrieval-based motion generation, we aim for MOST to inspire further advancements and deeper exploration within the text-to-motion field. Besides, while the generated motion sequences align with rare text prompts, some artifacts, such as footsliding, may occasionally occur. In the future, these issues can be mitigated by introducing additional constraints and enhancing physical simulations.


\section{Conclusion}

This study introduces MOST, a novel method for generating motion from rare text, addressing motion redundancy in prior approaches. MOST leverages common motion clips to guide less common ones through retrieval and generation stages. In retrieval, our temporal clip Banzhaf interaction enhances accuracy by capturing detailed text-motion clip connections, reducing redundancy. In generation, our motion prompt module integrates retrieved clips to generate motion aligned with rare text semantics, effectively utilizing key motion clips to tackle redundancy. Rigorous qualitative and quantitative experiments demonstrate MOST's superiority over existing methods, particularly in handling rare texts.

\bibliographystyle{IEEEtran}
\bibliography{main}

\end{document}